\documentclass{article}

\usepackage{arxiv}

\usepackage[utf8]{inputenc} % allow utf-8 input
\usepackage[T1]{fontenc}    % use 8-bit T1 fonts
\usepackage{hyperref}       % hyperlinks
\usepackage{url}            % simple URL typesetting
\usepackage{booktabs}       % professional-quality tables
\usepackage{amsfonts}       % blackboard math symbols
\usepackage{nicefrac}       % compact symbols for 1/2, etc.
\usepackage{microtype}      % microtypography
\usepackage{graphicx}
\usepackage{doi}

\title{Exploring the Potential of Large Language Models in Graph Generation}

\date{}

\usepackage{authblk}

\author[1]{%
	Yang Yao%
}
\author[1,*]{%
	Xin Wang%
}
\author[1]{%
    Zeyang Zhang%
}
\author[1]{%
	Yijian Qin%
}
\author[1]{%
	Ziwei Zhang%
}
\author[1]{%
    Xu Chu%
}
\author[1]{%
    Yuekui Yang%
}
\author[1,*]{%
	Wenwu Zhu%
}
\author[2]{%
	Hong Mei%
}
\affil[1]{Tsinghua University}
\affil[2]{Peking University}
\affil[*]{Corresponding authors.}

\usepackage{amsmath}
\usepackage{enumitem}
\usepackage{adjustbox}
\usepackage{multirow}
\usepackage{natbib}
\usepackage{subcaption}
\newcommand{\model}{\textbf{LLM4GraphGen} }
\newcommand{\modelnosp}{\textbf{LLM4GraphGen}}

\begin{document}
\maketitle

\begin{abstract}
  Large language models (LLMs) have achieved great success in many fields, and recent works have studied exploring LLMs for graph discriminative tasks such as node classification. 
  However, the abilities of LLMs for graph generation remain unexplored in the literature. Graph generation requires the LLM to generate graphs with given properties, which has valuable real-world applications such as drug discovery, while tends to be more challenging.
  In this paper, we propose LLM4GraphGen to explore the ability of LLMs for graph generation with systematical task designs and extensive experiments. 
  Specifically, we propose several tasks tailored with comprehensive experiments to address key questions regarding LLMs' understanding of different graph structure rules, their ability to capture structural type distributions, and their utilization of domain knowledge for property-based graph generation.
  Our evaluations demonstrate that LLMs, particularly GPT-4, exhibit preliminary abilities in graph generation tasks, including rule-based and distribution-based generation. We also observe that popular prompting methods, such as few-shot and chain-of-thought prompting, do not consistently enhance performance. Besides, LLMs show potential in generating molecules with specific properties. These findings may serve as foundations for designing good LLMs based models for graph generation and provide valuable insights and further research.
\end{abstract}

\section{Introduction}
Large language models (LLMs) have experienced remarkable success across various domains~\cite{zhao2023survey}. In comparison to their predecessors, LLMs possess a substantial number of parameters, enabling them to exhibit exceptional capabilities, notably being foundation models~\cite{bommasani2022opportunities}, in-context learning~\cite{GPT3,bubeck2023sparks}, chain-of-thought~\cite{COT}, {\it etc}. The application of LLMs has yielded impressive outcomes in numerous tasks, such as code generation~\cite{code1, code2}, drug discovery~\cite{drug1,drug2}, knowledge probing~\cite{know1,know2}, {\it etc}.

Despite not being explicitly designed for graph-structured data, exploring the potential of LLMs in comprehending and leveraging graph structures has attracted considerable attention recently~\cite{zhang2023graph}.
Notably, recent research endeavors have showcased promising results, demonstrating that LLMs can effectively address various discriminative tasks associated with graph data~\cite{wang2023can,ye2023natural,huang2023llms,tang2023graphgpt}, such as node classification, link prediction, and graph classification.
These studies highlight the inherent capacity of large language models to grasp the underlying structure of explicitly provided graph data, validating their potential in graph-related problem-solving.

On the other hand, graph generation has been a prominent and extensively studied research field in recent years~\cite{zhu2022survey}. It involves the generation of graphs with specific properties, and its significance extends to various domains, including drug discovery~\cite{you2018graph}.
Deep generative models have shown promising results in graph generation tasks, particularly in molecular generation. However,  current investigations of LLMs for graph-related tasks have primarily focused on discriminative tasks. Consequently, whether LLMs possess the abilities of graph generation remains unexplored in the literature.

\begin{figure*}[ht]
  \centering
  \includegraphics[width=\linewidth]{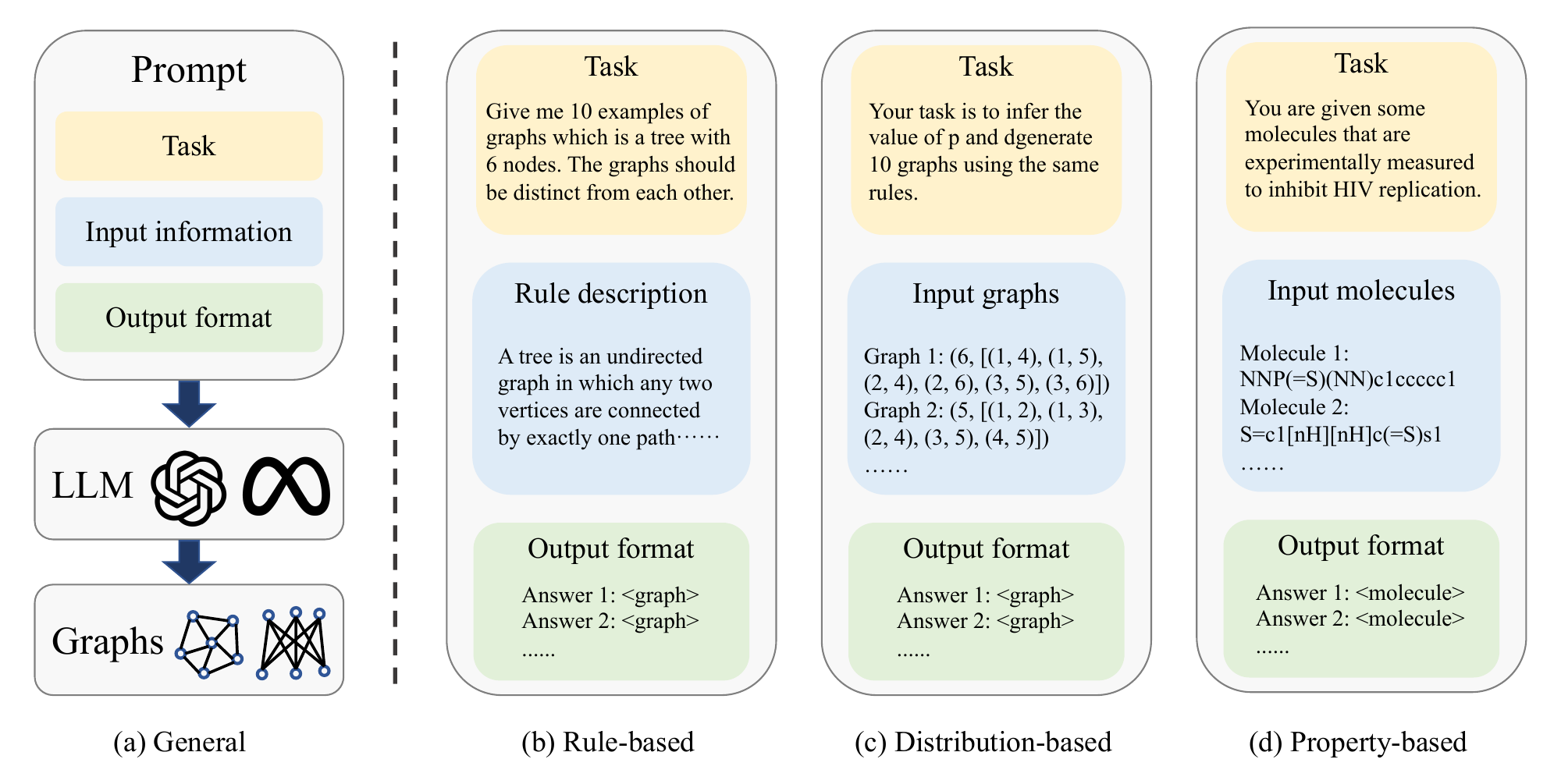}
  \caption{An overview of \modelnosp. Our proposed method designs a prompt tailored to each graph generation task, which is subsequently used as the input to the LLM to generate the desired graphs. Each prompt encompasses both the task description and the required output format. In the case of rule-based generation, the prompt contains the description of the rule. For distribution-based generation, a collection of graphs is provided to facilitate the LLM's learning of the underlying distribution. For property-based generation, a collection of molecules is included to enable the LLM to understand molecular properties.}
  \label{fig:framework}
\end{figure*}

In this work, we propose \model to explore the potential of LLMs for graph generation.
Specifically, we construct a pipeline where graph generation tasks are formulated as textual prompts and LLMs are required to output graphs in a specific format.
Within this pipeline, we explore three key questions of applying LLMs to graph generation tasks. \textbf{1) Whether the LLMs can understand the rules of different types of graph structures?} Understanding basic graph structure types through rules, {\it e.g.}, trees, cycles, regular graphs, and so on, is the foundation of graph generation. To answer this question, we benchmark LLMs for eight rule-based graph generation. Moreover, we explore the impact of important parameters and prompt construction methods for rule-based graph generation capabilities. \textbf{2) Whether LLMs can understand the distribution of different types of graph structures?}  We assess the ability of LLMs to understand the structural type distribution of given graphs and generate new graphs with the same distribution. \textbf{3) Whether the LLMs can understand domain knowledge of graph generation?} We explore LLMs in the task of property-based graph generation, particularly focusing on generating molecule structures with specific properties. This task necessitates a comprehensive understanding of both graph structures and the incorporation of domain knowledge.

Moreover, we have conducted extensive experiments and provided detailed results in this paper. By analyzing the results of the experiment, we derive several valuable observations. 
Our experiments demonstrate that GPT-4 has reasonably good abilities in graph generation, including rule-based and distribution-based generation. We also observe that some popular prompting methods, such as in-context learning and chain-of-thought, do not improve graph generation performance consistently. Moreover, LLMs show preliminary abilities in generating molecules with certain properties.
The conclusions drawn from our experiments provide reliable and informative insights that can guide future research and practical applications for graph generation.

The contributions of our work are summarized as:
\begin{itemize}[leftmargin = 0.5cm]
    \item We have explored the potential of using LLMs for graph generation. To the best of our knowledge, we are the first to study this valuable problem.
    \item We have designed comprehensive experiments to evaluate the graph generation ability of LLMs by proposing tasks with varying difficulty, including rule-based graph generation, distribution-based graph generation, and property-based graph generation.
    \item We extensively evaluate the graph generation capability of state-of-the-art LLMs, such as GPT-4, with diverse prompting methods under three metrics. Our experimental results reveal interesting and insightful observations that can inspire further research.
\end{itemize}

\section{\model}
In this section, we design three types of tasks to explore the graph generation ability of LLMs from multiple aspects, including rule-based tasks, distribution-based tasks, and property-based tasks.

\subsection{Rule-based graph generation}
\begin{figure}
  \centering
  \includegraphics[width=0.6\linewidth]{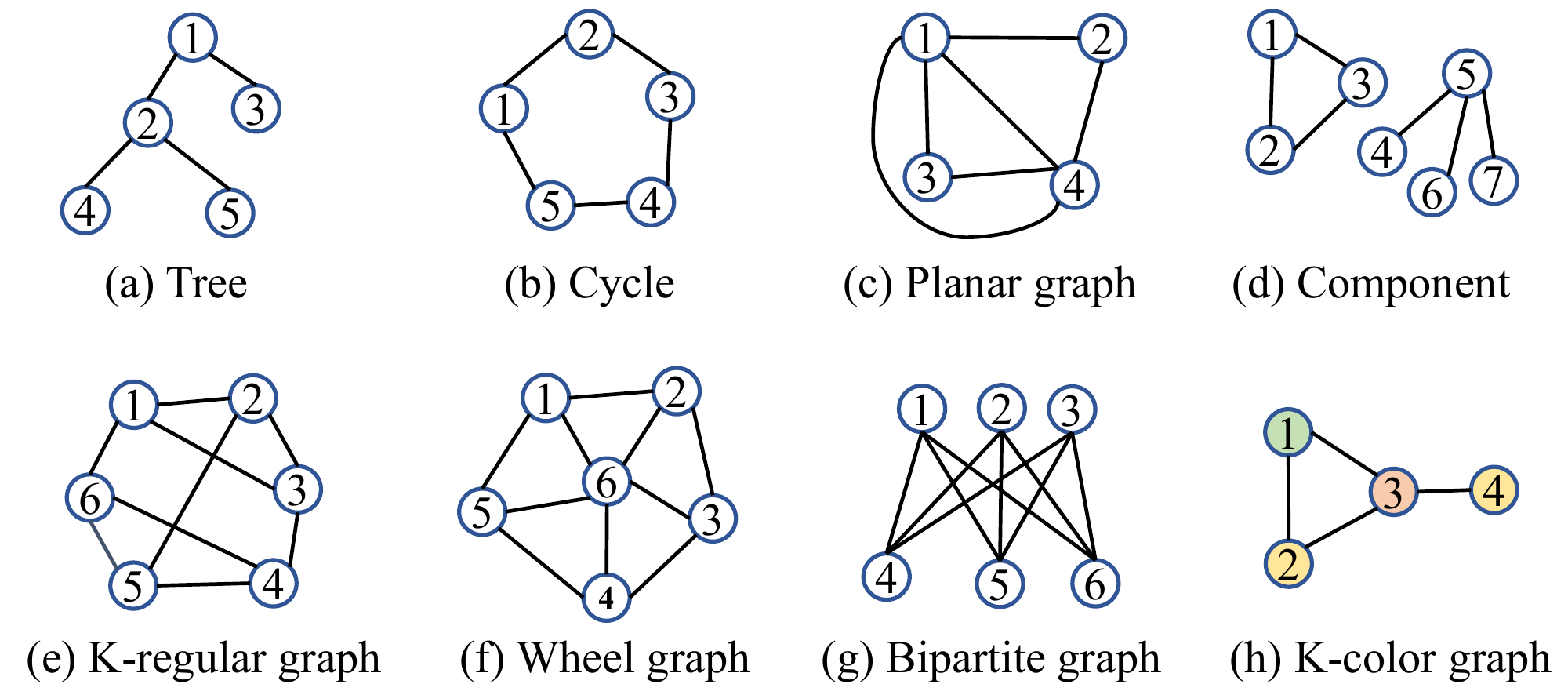}
  \caption{An illustration of graphs with regard to different rules.}
  \label{fig:rule}
\end{figure}
In this task, we employ rules in natural language to describe the structure of various graphs ({\it e.g.}, being a tree, or a cycle), and the model is evaluated by the adherence of its generated graphs to these rules. 
In order to assess the capacity of LLMs to comprehend and follow the instructions in graph generation, we formulate the following eight rules. Figure~\ref{fig:rule} shows the example graphs with regard to different rules.  
\begin{itemize}[leftmargin = 0.5cm]
    \item \textbf{Trees}: The generated graph should be a tree with the specified number of nodes, {\it i.e.}, an undirected graph where there is exactly one path between any two nodes.
    \item \textbf{Cycles}: The generated graph should contain a cycle with the specified number of nodes, and should not contain any other nodes or edges. 
    \item \textbf{Planar graphs}: The generated graph should be a planar graph with the specified number of nodes and edges, {\it i.e.}, there exists a way to draw the graph on a plane without edge crossings.
    \item \textbf{Components}: The generated graph should have a specified number of connected components, {\it i.e.}, there must be a specified number of connected subgraphs with no edges among them.
    \item \textbf{$k$-regular graphs}: Every node of the generated graph should have the same number of neighbors, {\it i.e.}, the degree of every node is $k$.
    \item \textbf{Wheel graphs}: The generated graph should be formed by connecting a single node to all nodes of a cycle.
    \item \textbf{Bipartite graphs}: The generated graph should have nodes that are divided into two disjoint and independent sets $U$ and $V$ with specified sizes.
    \item \textbf{$k$-color graphs}: The generated graph should be $k$-colorable, {\it i.e.}, each node is assigned a color, and two adjacent nodes do not have the same color.
\end{itemize}

\paragraph{Prompt Design} We adopt the following four types of prompts for rule-based graph generation: 
\begin{itemize}[leftmargin = 0.5cm]
    \item \textbf{Zero-shot}: The prompt contains the relevant information about the rules, as well as a specification of the output format. The model is then asked to generate graphs using the given rules.
    \item \textbf{Few-shot}: In addition to the zero-shot prompt, the model is given several graph examples that follow the given rules. The edges of the graphs are sorted by the node ID to facilitate the model understanding.
    \item \textbf{Zero-shot+CoT}/\textbf{Few-shot+CoT}: In addition to the zero-shot prompt and the few-shot prompt, the model is asked to give the answer step by step.
\end{itemize}

\subsection{Distribution-based graph generation}
\label{dbgg}

Learning distributions from the given graphs and then generating new graphs based on the distributions is a necessary ability for graph generation~\cite{zhu2022survey}.
To this end, we propose distribution-based graph generation tasks, which define a target graph distribution and the generated set of graphs $\{G_i\}$ from the target distribution as the input to the model.
The model should infer the parameters of the target distribution from the given graphs, and generate new graphs from the same distribution. We design several subtasks with increasing difficulties, including the generation of tree-or-cycles, union of components, and motifs.

\begin{figure}[t]
  \centering
  \includegraphics[width=0.5\linewidth]{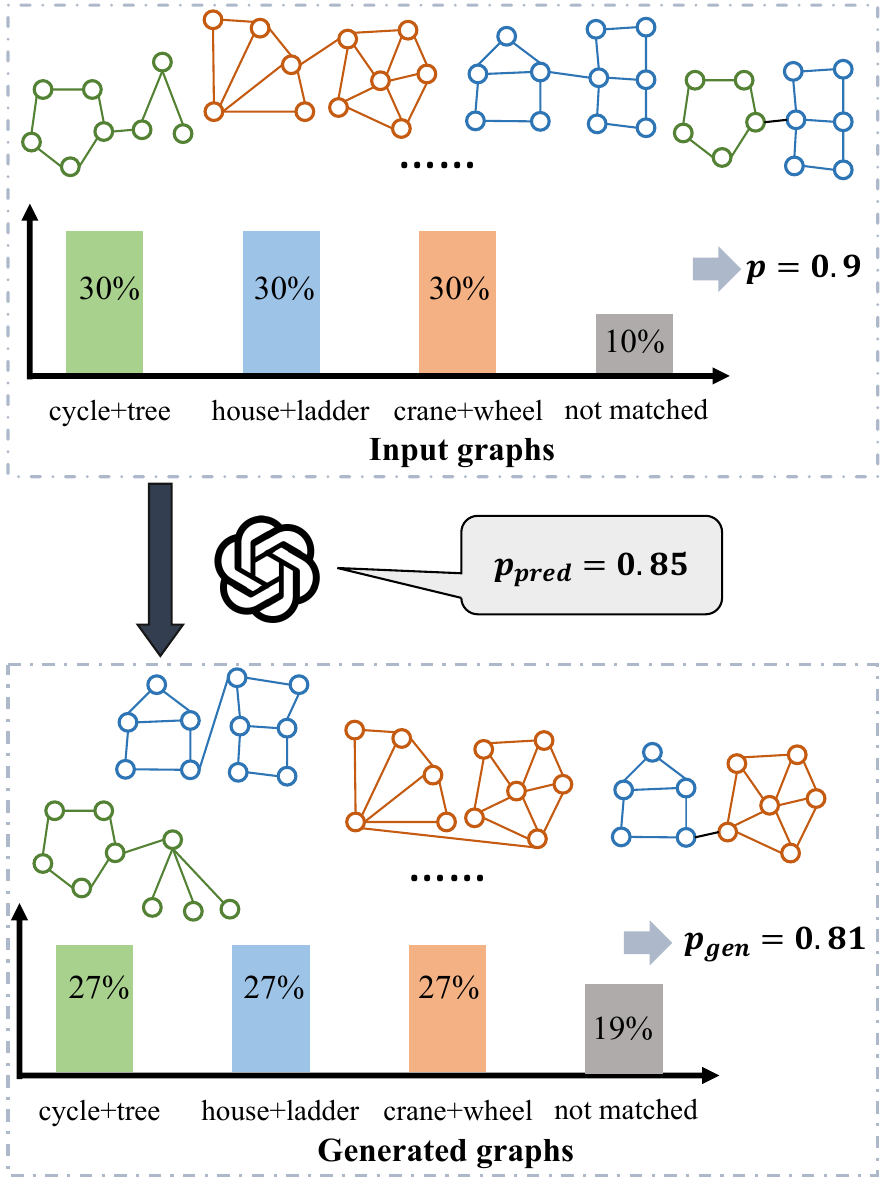}
  \caption{An illustration of distribution-based graph generation. }
  \label{fig:newdist}
\end{figure}

\paragraph{Trees or cycles}
In this subtask, the target distribution is defined as a mixture of trees and cycles, where each graph has a probability of $p$ to be a tree and $1-p$ to be a cycle,
\begin{equation}
    P(G_i \text{ is a tree}) = p, \quad P(G_i \text{ is a cycle}) = 1-p.
\end{equation}
Specifically, the model is given 10 graphs sampled from the target distribution and is asked to infer the value of $p$ from the graphs and then generate 10 new graphs that follow the same distribution.
We evaluate the task with different settings of $p$, and all input graphs have a random number of nodes ranging from 5 to 7.

\paragraph{Union of components}
In this subtask, each graph from the target distribution is the union of two connected components, with each component being either a tree or a cycle. There is a probability of $p$ that the two components belong to the same kind and $1-p$ that they belong to different kinds, {\it i.e.}, $G_i = G_i^1 \cup G_i^2$.
The model is asked to generate graphs with two component types specified by rules, and similarly, the input to the model is 10 random graphs drawn from the target distribution. The model is expected to infer the value of $p$ and generate 10 new graphs from the target distribution. Each component has a random number of 5 to 7 nodes. This task is more complex and evaluates the ability of graph generation in scenarios with multiple correlated factors such as the component types.

\paragraph{Motif}
In this subtask, each graph from the target distribution consists of a base graph and a motif graph with an inter-connected edge, where there are three kinds of base graphs (trees, ladders, wheels), and three motif graphs (cycles, houses, cranes). Inspired by the setting of Spurious-Motif~\cite{gnnexplainer,dir,Grace}, we construct the distribution as follows. 
For each graph $G_i$, the base graph type $S_i$ is chosen from a uniform distribution over three kinds of base graphs, and the motif graph $C_i$ with the same label is chosen with probability $p$, while the other two motif graphs are chosen with probability $(1-p)/2$ respectively, {\it i.e.},
\begin{equation}
    P(S_i, C_i) = \begin{cases}
        \frac{1}{3} p & S_i = C_i, i\in \{0,1,2\} \\
        \frac{1}{6} (1-p) & S_i \neq C_i, i\in \{0,1,2\}
    \end{cases}.
\end{equation}
We generate 10 random graphs from the target distribution and ask the model to infer the value of $p$, and generate 10 new graphs from the same distribution.
The main purpose of this task is to explore the abilities of LLMs to understand graph distributions under more complex factors and generate graphs based on inferred distributions.

\paragraph{Prompt Design} In zero-shot prompt, we introduce the graph generation task and the target distribution, and provide a set of graphs sampled from the target distribution as the input. Then, we ask the model to infer the value of $p$ and generate new graphs. Similarly, we add exemplars in few-shot prompt, and step-by-step thinking in CoT prompt.

\subsection{Property-based graph generation}
\begin{figure}[t]
  \centering
  \includegraphics[width=0.5\linewidth]{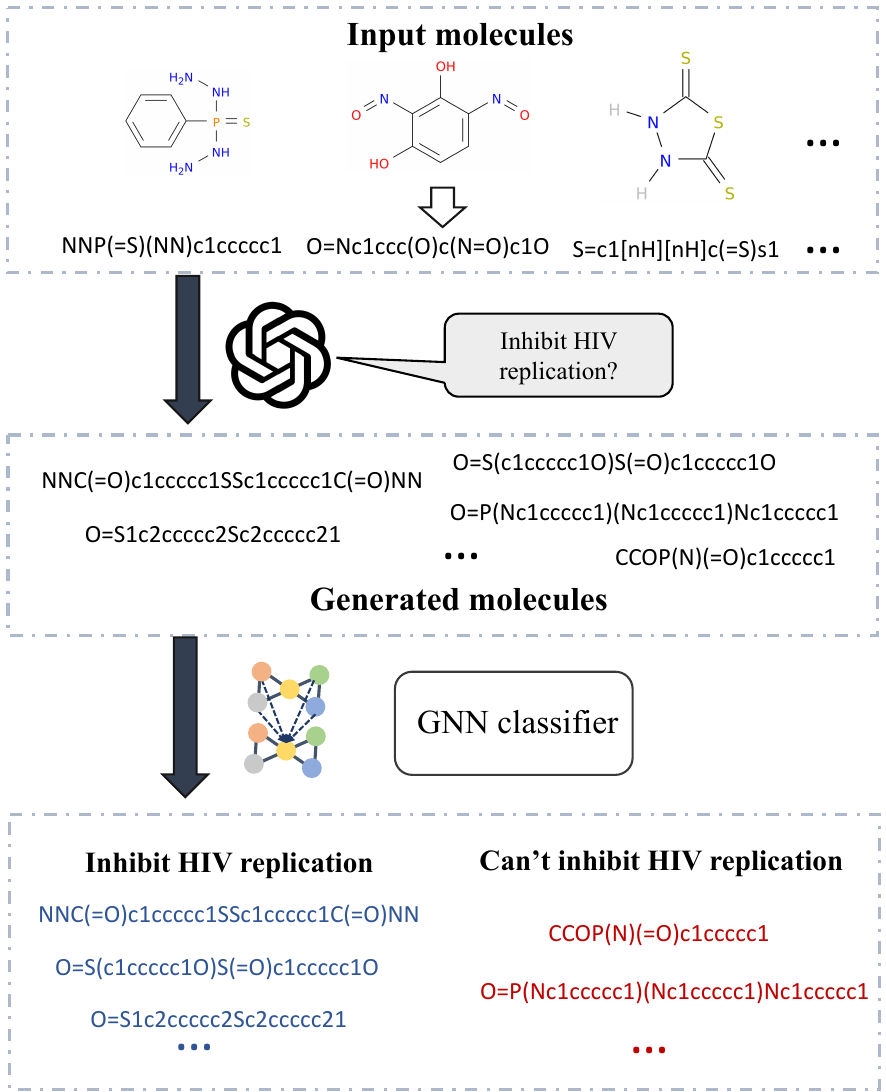}
  \caption{An illustration of property-based graph generation. }
  \label{fig:property}
\end{figure}

Generating graphs with certain properties is important in real-world applications, {\it e.g.}, the properties, in the field of drug discovery, such as inhibiting HIV replication,  blood-brain barrier penetration, and toxicity to the human body, are important for the development of new drugs~\cite{ggsurvey, zhu2022survey}. Though LLMs have learned expert knowledge like chemistry and medicine through large-scale textual data, it remains unknown whether they can leverage this information to directly generate graphs with desired properties.

In this task, we evaluate LLMs' abilities to generate graphs with given properties by adopting the molecular property prediction dataset OGBG-MolHIV~\cite{hu2020open}. We choose the molecules labeled to inhibit HIV replication as the input graphs and ask the model to generate new molecules with properties similar to these graphs. Following the common practice in the field of chemistry, we adopt the SMILES~\cite{weininger1988smiles} format to represent the molecule graphs.

To evaluate the performance of molecule generation, we measure the fraction of generated molecules that have the same properties as the input molecules. 
Since it would be infeasible to check the properties manually using experiments or expert knowledge, we use a GNN classifier trained on the OGBG-MolHIV dataset to assist in the evaluation of molecule generation.
Specifically, let $G_1, G_2, \ldots, G_n$ be the generated graphs that represent molecules, $C_M(G) \in [0,1]$ be the prediction given by the classifier, {\it e.g.}, $C_M(G) = 0.5$ means the molecule $G$ is predicted to have the desired property with 50\% probability. Since the classifier may not be exactly accurate, we calculate the rectified predictions $C(G)$ by taking into consideration the dataset priors as well as the classifier accuracy in the original dataset. 

\paragraph{Prompt Design} We design a few-shot prompt that the LLM is given a description of the desired property and a collection of molecules that have the property. Then, the model is asked to generate new molecules with the same property. For the CoT prompt, the model is also asked to provide a step-by-step explanation of the answer.

\begin{table*}
    \centering
  \caption{The valid rate for rule-based graph generation with GPT-4. The metric measures the fraction of generated graphs that are valid under the specified rules. Values after $\pm$ denote standard errors.}
  \label{tab:rule-total}
  \adjustbox{max width=\linewidth}{
  \begin{tabular}{cccccccccc}
    \toprule
    Prompt&Trees&Cycles& Components &Planar &$k$-regular&Wheel&Bipartite&$k$-color\\
    \midrule
    Zero-shot & $100.0 \pm 0.0$ & $91.3 \pm 3.3$ & $30.4 \pm 5.1$ & $47.3 \pm 4.2$ & $64.0 \pm 6.8$ & $13.0 \pm 5.3$ & $60.3 \pm 7.4$ & $50.3 \pm 5.5$ \\
    Few-shot & $98.0 \pm 0.9$ & $85.0 \pm 3.3$ & $63.2 \pm 5.3$ & $4.3 \pm 1.3$ & $86.1 \pm 3.1$ & $88.8 \pm 7.4$ & $57.1 \pm 8.6$ & $62.3 \pm 5.1$ \\
    Zero-shot+CoT & $100.0 \pm 0.0$ & $86.9 \pm 3.6$ & $38.0 \pm 5.1$ & $53.3 \pm 6.0$ & $82.7 \pm 8.6$ & $92.3 \pm 4.7$ & $92.7 \pm 4.4$ & $43.2 \pm 4.9$ \\
    Few-shot+CoT & $97.6 \pm 1.7$ & $97.0 \pm 1.9$ & $40.0 \pm 6.7$ & $20.0 \pm 4.3$ & $91.5 \pm 1.6$ & $90.7 \pm 5.1$ & $98.2 \pm 1.8$ & $58.5 \pm 5.9$ \\
    \bottomrule
  \end{tabular}
  }
\end{table*}

\begin{table*}
  \centering
  \caption{The unique rate for rule-based graph generation with GPT-4. The metric measures the fraction of valid graphs that are unique under the specified rules.}
  \label{tab:rule-total-unique}
  \adjustbox{max width=\linewidth}{
  \begin{tabular}{cccccccccc}
    \toprule
    Prompt&Trees&Cycles& Components &Planar &$k$-regular&Wheel&Bipartite&$k$-color\\
    \midrule
    Zero-shot  & $98.6 \pm 0.9$ & $88.3 \pm 1.6$ & $91.9 \pm 3.0$ & $99.7 \pm 0.3$ & $100.0 \pm 0.0$ & $98.7 \pm 0.8$ & $44.3 \pm 7.0$ & $98.3 \pm 1.6$ \\
    Few-shot  & $99.3 \pm 0.5$ & $92.3 \pm 1.3$ & $97.7 \pm 1.1$ & $96.8 \pm 2.5$ & $100.0 \pm 0.0$ & $98.8 \pm 1.1$ & $50.0 \pm 8.2$ & $98.5 \pm 1.5$ \\
    Zero-shot+CoT  & $100.0 \pm 0.0$ & $89.0 \pm 4.2$ & $98.5 \pm 0.8$ & $98.6 \pm 0.8$ & $100.0 \pm 0.0$ & $85.0 \pm 5.7$ & $16.9 \pm 4.2$ & $98.6 \pm 1.0$ \\
    Few-shot+CoT & $99.7 \pm 0.3$ & $83.0 \pm 5.1$ & $96.3 \pm 1.5$ & $98.0 \pm 1.4$ & $100.0 \pm 0.0$ & $88.3 \pm 5.1$ & $10.7 \pm 0.7$ & $96.3 \pm 2.6$ \\
    \bottomrule
  \end{tabular}
  }
\end{table*}

\section{Experimental Results and Analyses}
In this section, we conducted extensive experiments on LLMs for our proposed graph generation tasks. Through analyzing the results, we derive valuable observations to benefit practical applications.

\subsection{Rule-based graph generation}
We explore the graph generation abilities of several representative LLMs in this part, including GPT-4, GPT-3.5, and LLama2-13B. We use the following three metrics to evaluate the graph generation quality:
\begin{itemize}[leftmargin = 0.5cm]
    \item \textbf{Valid rate}: The proportion of generated graphs that match the rules.
    \item \textbf{Novel rate}: The proportion of generated graphs that is different from the given example graph.
    \item \textbf{Unique rate}: The proportion of generated graphs that are not identical.
\end{itemize}
Since GPT-4 is the strongest LLM currently available, we perform our main experiments with GPT-4. The valid rate and unique rate for various rules and prompts are listed in Table \ref{tab:rule-total} and Table \ref{tab:rule-total-unique} respectively.
It is worth noting that by designing appropriate prompts, the novel rate of each experiment is $100\%$ or close to $100\%$ for GPT-4.
More details about the setups and results can be found in the appendix.

\textbf{Observation 1}: GPT-4 has reasonably good abilities for rule-based graph generation.   

Table~\ref{tab:rule-total} demonstrates that GPT-4 has the ability to generate graphs according to rules in general. Specifically, for simple rules such as Trees and Cycles, GPT-4 achieves good generation quality. But the generation quality is not good enough for other rules. 
A possible reason is that the topological structures of trees and cycles are relatively simple. LLMs can quickly come up with their generation algorithms according to the massive training data. 
For more complex rules like $k$-regular, wheel, and bipartite, LLMs cannot achieve satisfactory results with zero-shot prompt. Nevertheless, adjusting prompts such as few-shot and chain-of-thought can improve the generation quality.
For components, planar graphs, and $k$-color graphs, it is difficult for LLMs to find the accurate generation method from rule descriptions, even in conditions giving few-shot examples and CoT prompt, resulting in poor generation quality.

In order to explore the graph generation effects of different LLMs, we conducted experiments on multiple LLMs and generated small-size graphs, as shown in Table~\ref{tab:rule-llm}. It can be found that GPT-4 has good generation quality for all three rules. GPT-3.5 can generate some legitimate cycles, but its performance is poor for more complex rules. LLama2 can only generate very few graphs that comply with rules given examples.

\begin{table}
  \centering
  \caption{A comparison of rule-based graph generation using different LLMs. Reported values are valid rates.}
  \label{tab:rule-llm}
  \adjustbox{max width=\linewidth}{
  \begin{tabular}{ccccc}
    \toprule
    Model&Prompt&Cycle&$k$-regular&$k$-color\\
    \midrule
    \multirow{4}{*}{GPT-4} & Zero-shot & $84.7 \pm 5.9$ & $74.7 \pm 9.9$ & $56.0 \pm 4.5$ \\
     & Few-shot & $73.3 \pm 7.1$ & $90.0 \pm 2.9$ & $76.4 \pm 5.3$ \\
     & Zero-shot+CoT & $95.3 \pm 3.2$ & $97.1 \pm 1.2$ & $62.0 \pm 6.1$ \\
     & Few-shot+CoT & $96.0 \pm 2.8$ & $87.9 \pm 3.4$ & $80.7 \pm 3.2$ \\
    \midrule
    \multirow{4}{*}{GPT-3.5} & Zero-shot & $84.7 \pm 5.6$ & $6.7 \pm 2.4$ & $0.0 \pm 0.0$ \\
     & Few-shot & $22.7 \pm 7.4$ & $8.7 \pm 3.6$ & $29.3 \pm 7.0$ \\
     & Zero-shot+CoT & $56.7 \pm 8.0$ & $2.7 \pm 2.6$ & $14.3 \pm 8.4$ \\
     & Few-shot+CoT & $57.5 \pm 12.1$ & $9.0 \pm 4.6$ & $34.3 \pm 8.0$ \\
    \midrule
    \multirow{4}{*}{LLama2} & Zero-shot & $0.7 \pm 0.7$ & $0.0 \pm 0.0$ & $0.0 \pm 0.0$ \\
     & Few-shot & $31.1 \pm 7.8$ & $17.2 \pm 5.0$ & $13.0 \pm 3.7$ \\
     & Zero-shot+CoT & $1.1 \pm 1.1$ & $0.0 \pm 0.0$ & $0.0 \pm 0.0$ \\
     & Few-shot+CoT & $16.2 \pm 5.2$ & $8.0 \pm 3.1$ & $13.3 \pm 3.7$ \\
    \bottomrule
  \end{tabular}
  }
\end{table}

\subsubsection{The impact of prompt on graph generation}
It is known that prompts have a significant impact on the performance of LLMs. In this section, we compare the performance of different prompts for rule-based graph generation with GPT-4. The valid rate and unique rate for different prompts are listed in Table \ref{tab:rule-total} and Table \ref{tab:rule-total-unique} respectively.
We have the following observations.

\textbf{Observation 2}: Providing examples has an inconsistent impact on LLMs in generating different types of graphs. 

Giving examples can have an impact on generation quality, but not all have a positive impact. 
As shown in Table~\ref{tab:rule-total}, for trees, the valid rate of few-shot prompt is worse than zero-shot prompt, which is because adding examples may disrupt the understanding of LLM for rule descriptions.
For planar graphs, giving examples greatly reduces the valid rate of the generated graph, as it is difficult to generalize the properties of the planar graph from the examples, and instead may lead to some opposite rules.
For wheel graphs, the valid rate of few-shot is much better than zero-shot because the generation of a wheel can actually be divided into two steps: first, generate a cycle, and then add a new node to connect it to each previous node. LLM can learn about the existence of the cycle from the examples, which strengthens its understanding of the rules.

\textbf{Observation 3}: CoT prompt has diverse impacts on different evaluation metrics for graph generation. 

CoT prompt can have an impact on generation quality, but not all have a positive impact. Making LLM think step by step can help it better decompose tasks. For example, as mentioned earlier, generating a wheel graph can be divided into two steps. 
Generating a bipartite graph can also be divided into two steps: firstly, the nodes are divided into two subsets, and then the nodes in the two subsets are connected. 
However, although the valid rate has increased, its unique rate has greatly decreased because, after step-by-step thinking, it always tends to divide nodes into the same subset.

\begin{table*}
  \caption{The comparison of different graph sizes for rule-based graph generation with GPT-4. ``Valid'' denotes fractions of generated graphs that are valid under specific rules, and ``Unique'' denotes fractions of generated graphs that are not duplicates.}
  \label{tab:rule-size}
  \centering
  \adjustbox{max width=\linewidth}{
  \begin{tabular}{ccccccccc}
    \toprule
    \multirow{2}{*}{Size} &\multirow{2}{*}{Prompt}& \multicolumn{2}{c}{Cycles}& \multicolumn{2}{c}{$k$-regular}& \multicolumn{2}{c}{$k$-color}\\
    &&Valid&Unique&Valid&Unique&Valid&Unique\\
    \midrule
    \multirow{4}{*}{Small} & Zero-shot & $84.7 \pm 5.9$ & $90.0 \pm 3.1$ & $74.7 \pm 9.9$ & $100.0 \pm 0.0$ & $56.0 \pm 4.5$ & $100.0 \pm 0.0$ \\
& Few-shot & $73.3 \pm 7.1$ & $96.0 \pm 1.8$ & $90.0 \pm 2.9$ & $100.0 \pm 0.0$ & $76.4 \pm 5.3$ & $100.0 \pm 0.0$ \\
& Zero-shot+CoT & $95.3 \pm 3.2$ & $84.7 \pm 4.8$ & $97.1 \pm 1.2$ & $100.0 \pm 0.0$ & $62.0 \pm 6.1$ & $96.7 \pm 3.2$ \\
& Few-shot+CoT & $96.0 \pm 2.8$ & $94.7 \pm 2.1$ & $87.9 \pm 3.4$ & $100.0 \pm 0.0$ & $80.7 \pm 3.2$ & $100.0 \pm 0.0$ \\
\midrule
\multirow{4}{*}{Medium} & Zero-shot & $91.3 \pm 3.3$ & $88.3 \pm 1.6$ & $64.0 \pm 6.8$ & $100.0 \pm 0.0$ & $50.3 \pm 5.5$ & $98.3 \pm 1.6$ \\
& Few-shot & $85.0 \pm 3.3$ & $92.3 \pm 1.3$ & $86.1 \pm 3.1$ & $100.0 \pm 0.0$ & $62.3 \pm 5.1$ & $98.5 \pm 1.5$ \\
& Zero-shot+CoT & $86.9 \pm 3.6$ & $89.0 \pm 4.2$ & $82.7 \pm 8.6$ & $100.0 \pm 0.0$ & $43.2 \pm 4.9$ & $98.6 \pm 1.0$ \\
& Few-shot+CoT & $97.0 \pm 1.9$ & $83.0 \pm 5.1$ & $91.5 \pm 1.6$ & $100.0 \pm 0.0$ & $58.5 \pm 5.9$ & $96.3 \pm 2.6$ \\
\midrule
\multirow{4}{*}{Large} & Zero-shot & $96.0 \pm 1.6$ & $84.0 \pm 4.4$ & $61.3 \pm 10.6$ & $100.0 \pm 0.0$ & $44.3 \pm 6.8$ & $97.9 \pm 1.5$ \\
& Few-shot & $82.0 \pm 6.7$ & $96.7 \pm 2.0$ & $70.0 \pm 7.4$ & $100.0 \pm 0.0$ & $64.0 \pm 8.8$ & $74.0 \pm 13.7$ \\
& Zero-shot+CoT & $100.0 \pm 0.0$ & $85.6 \pm 9.0$ & $70.0 \pm 21.2$ & $100.0 \pm 0.0$ & $58.9 \pm 6.4$ & $95.6 \pm 4.2$ \\
& Few-shot+CoT & $95.0 \pm 3.6$ & $85.7 \pm 6.2$ & $82.5 \pm 7.1$ & $100.0 \pm 0.0$ & $53.3 \pm 10.8$ & $89.2 \pm 5.5$ \\
    
    \bottomrule
  \end{tabular}
  }
\end{table*}

\subsubsection{The impact of parameters on graph generation}

Graph generation using LLMs can be sensitive to the choice of parameters.
In order to explore the impact of various parameters on graph generation, we construct the ablation study.
We compare the performance of graph generation when using different graph sizes in Table \ref{tab:rule-size}. The ``Medium'' size in the experiment is the same as the size in the main experiment. More details of the experiment can be found in the appendix.
Through the analysis of the generated results, we have reached the following conclusion:

\textbf{Observation 4}: As the graph size increases, the performance of LLM in graph generation decreases for most rules, except for simple cases such as cycles.

Table~\ref{tab:rule-size} shows the valid rate and unique rate when using LLM to generate graphs of different sizes for three rules. In experience, the larger the graph generated, the poorer the generation quality, but this is not the case for cycles. This may be because the cycle is too simple. 

In addition, we can also observe from Table~\ref{tab:rule-size} that the larger the size, the smaller the unique rate, indicating that LLM tends to generate graphs from smaller sets. This is counterintuitive because the larger the size, the larger the set of graphs that satisfy the rules. This indicates that LLM can handle a certain amount of information, and the larger the size of the graph, the smaller the set of graphs.

\subsection{Distribution-based graph generation}
We use GPT-4 for graph generation tasks, and there are two metrics for evaluating the quality of generation:
\begin{itemize}[leftmargin=0.5cm]
    \item $p_{\text{pred}}$: The value of $p$ represents the distribution of a set of graphs. $p_{\text{pred}}$ is the value of $p$ predicted by the LLM using the set of input graphs.
    \item $p_{\text{gen}}$: The value of $p$ calculated by the generated graphs. 
\end{itemize}

We adopt three experiment settings (refer to Section~\ref{dbgg} for more details) with increasing difficulty.
The purpose of the experiments we designed is as follows:
\begin{itemize}[leftmargin=0.5cm]
    \item \textbf{Trees or cycles}: Exploring the distribution of graphs.
    \item \textbf{Union of components}: Exploring the distribution of subgraph combinations within a graph.
    \item \textbf{Motif}: Exploring the distribution of subgraph combinations within the graph for more complex situations.
\end{itemize}

\begin{figure*}[ht]
\centering
\subcaptionbox{$p_{\text{pred}}$ on ``Trees or cycles'' task}{\includegraphics[width=0.31\linewidth]{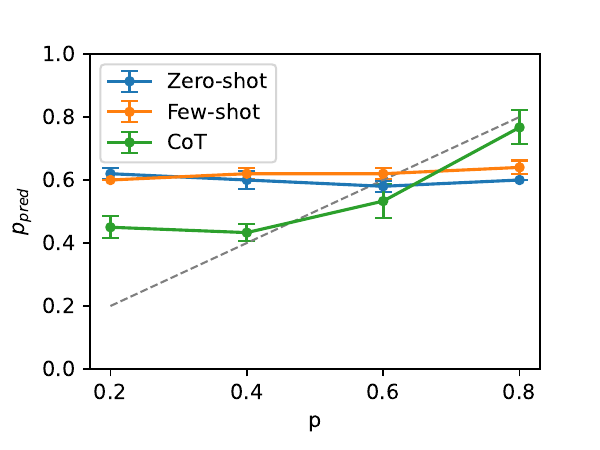}}
\subcaptionbox{$p_{\text{pred}}$ on ``Union of components'' task}{\includegraphics[width=0.31\linewidth]{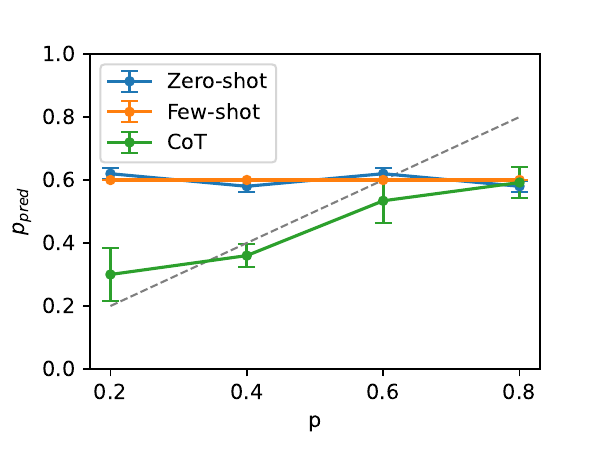}}
\subcaptionbox{$p_{\text{pred}}$ on ``Motif'' task}{\includegraphics[width=0.31\linewidth]{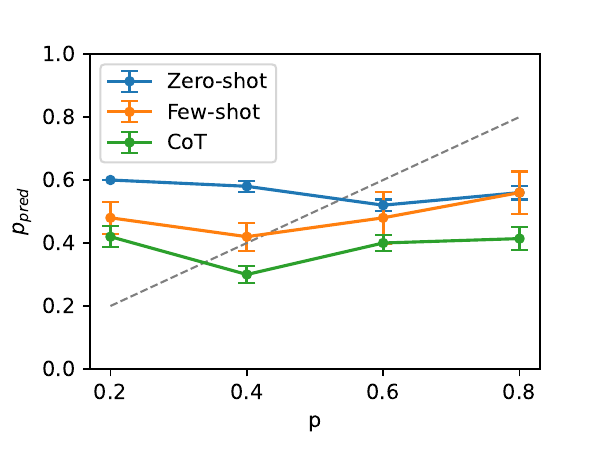}}
\subcaptionbox{$p_{\text{gen}}$ on ``Trees or cycles'' task}{\includegraphics[width=0.31\linewidth]{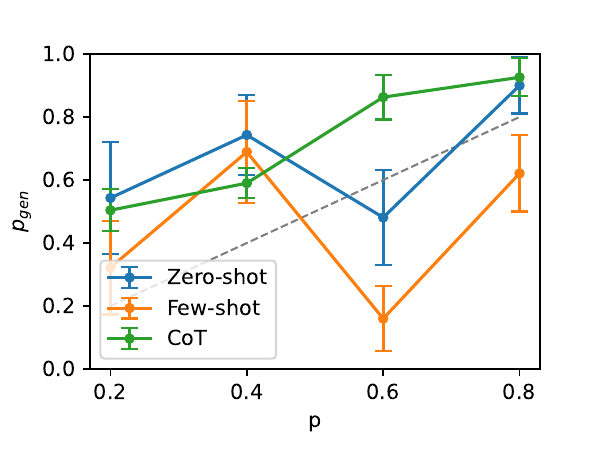}}
\subcaptionbox{$p_{\text{gen}}$ on ``Union of components'' task}{\includegraphics[width=0.31\linewidth]{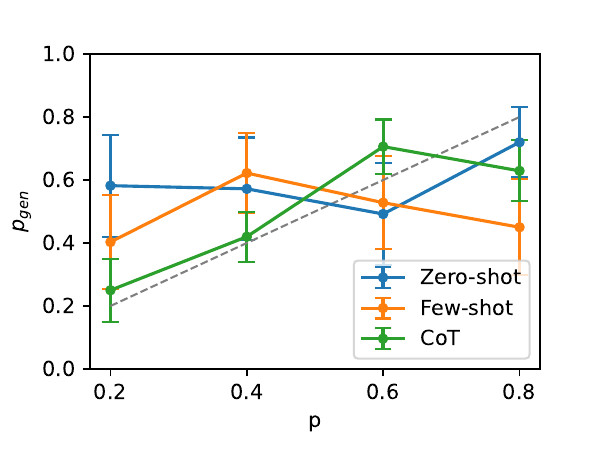}}
\subcaptionbox{$p_{\text{gen}}$ on ``Motif'' task}{\includegraphics[width=0.31\linewidth]{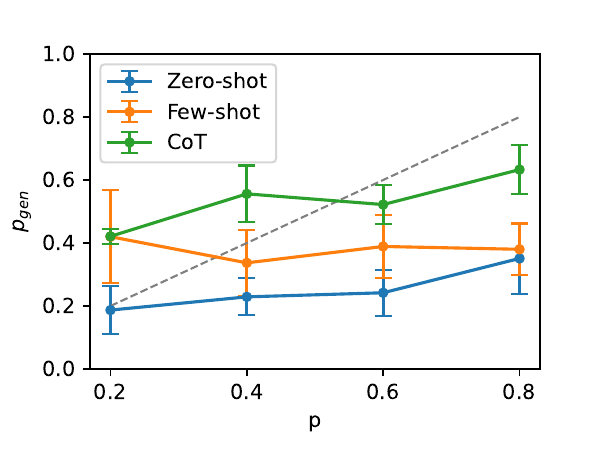}}
\caption{Performance of GPT-4 on distribution-based graph generation. $p$ represents the parameter of the distribution where the graphs are sampled. $p_{\text{pred}}$ is the value of $p$ predicted by LLM for the input graphs. $p_{\text{gen}}$ is the value of $p$ calculated by the generated graphs. In this task, the performance of LLMs is better when $p_{\text{pred}}$ and $p_{\text{gen}}$ are closer to the ground-truth parameter $p$. }
  \label{fig:dist-total}
\end{figure*}

The previous experiments have demonstrated that GPT-4 can generate trees and cycles with high accuracy. Therefore, our designed experiments on ``Trees or cycles'' can eliminate the impact of generation quality as much as possible.
In addition, we also constructed a set of experiments to demonstrate the ability of GPT-4 to generate two-component graphs with trees and cycles as subgraphs, as shown in Table~\ref{tab:distr-o}. This indicates that our constructed ``Union of components'' experiment can eliminate the impact of generation quality as much as possible.
The results can be seen in Figure \ref{fig:dist-total}, which provides the $p_{\text{pred}}$ and $p_{\text{gen}}$ values for the three tasks.

\textbf{Observation 5}: LLMs can understand and generate graphs with simple distributions, but perform poorly in complex situations.

Figure~\ref{fig:dist-total} shows the performance of GPT-4 on different tasks. From the figure, for ``Trees or cycles'' and ``Union of components'', we can see that although the values of $p_{\text{pred}}$, $p_{\text{gen}}$, and $p$ cannot be completely consistent when using CoT prompt, the curve trends are similar. Since we generate the graph based on probability rather than strictly proportional, this error is reasonable.
However, for ``Motif'' tasks, none of the three prompts learned anything useful, and their $p_{\text{pred}}$ always fluctuated around 0.5.

\textbf{Observation 6}: Detailed examples and CoT are helpful for distribution-based graph generation. 

From Figure~\ref{fig:dist-total}, it can be seen that for the first two tasks, the $p_{\text{pred}}$ and $p_{\text{gen}}$ generated by CoT are relatively accurate, while zero-shot and few-shot have hardly learned anything. This indicates that providing only examples is not enough, and it is necessary to demonstrate in detail how to estimate the distribution from the given graph and generate a new set of graphs based on the given p-value for GPT-4.

\begin{table}
\centering
  \caption{The valid rate of two-component graph generation.}
  \label{tab:distr-o}
    \adjustbox{max width=\linewidth}{
  \begin{tabular}{cccc}
    \toprule
    Prompt&Tree+Tree&Cycle+Cycle&Tree+Cycle\\
    \midrule
    Zero-shot &	$43.0 \pm 8.3$ &$39.0 \pm 6.1$&$38.0 \pm 11.9$\\
    Few-shot & $81.0 \pm 6.2$&$13.0 \pm 6.5$&$48.0 \pm 11.8$\\
    CoT &$100.0 \pm 0.0$&$100.0 \pm 0.0$&$89.0 \pm 4.4$\\
    \bottomrule
  \end{tabular}
  }
\end{table}

\subsection{Property-based graph generation}
We use GPT-4 for graph generation, and there are four metrics for evaluating the quality of graph generation:
\begin{itemize}
    \item \textbf{$C_M(G)$}: the classifier's predicted probability of having the desired properties for the generated molecules.
    \item \textbf{$C(G)$}: the rectified probability of having the desired properties for the generated molecules.
    \item \textbf{Novel rate}: the fraction of generated molecules that are distinct from the input molecules.
    \item \textbf{Unique rate}: the fraction of generated molecules that are not duplicates.
\end{itemize}
From Table \ref{tab:know}, we have the following observation.

\begin{table}
\centering
  \caption{Results of property-based graph generation. $C_M(G)$ is the classifier's predicted probability of having the desired properties for the generated molecules, while $C(G)$ is the rectified probability. ``Novel'' denotes generated molecules that are not the same as the input molecules, while ``Unique'' denotes molecules that are not duplicated with other generated molecules.}
  \label{tab:know}
  \adjustbox{max width=\linewidth}{
  \begin{tabular}{ccccc}
    \toprule
   Prompt&  $C_M(G)$ & $C(G)$ & Novel & Unique \\
   \midrule
Few-shot &$26.4 \pm 7.5$& $ 34.8 \pm 16.5$ & $79.1 \pm 10.9$ &$ 91.8 \pm 6.1$ \\
Few-shot+CoT &$32.7 \pm 4.7$&$ 48.8 \pm 10.2$ & $65.5 \pm 10.9$ &$ 92.7 \pm 6.0$ \\
    \bottomrule
  \end{tabular}
  }
\end{table}

\textbf{Observation 7}: LLMs show preliminary abilities in generating molecules with certain properties.

As shown in Table~\ref{tab:know}, both $C(G)$ and $C_M(G)$ are greater than 0, indicating that in the classifier's view, some of the generated molecules have specified properties. Although some molecules are duplicates of known molecules, there are still some new molecules that have this property. 
And it can be found that the generated molecules have relatively high levels of novel rate and unique rate.

In addition, we observe that both $C(G)$ and $C_M(G)$ increase when using CoT prompt, which indicating that it's beneficial for LLM to think step by step, so that the reason abilities of LLM can be activated in these complex graph generation scenarios.

\section{Related Work}

We review related work on large language model for graphs and graph generation respectively.

\subsection{Large Language Model for Graphs}

Recently, there have been several works about utilizing large language models (LLMs) for solving problems on graphs. \citet{he2023explanations} propose to utilize LLMs to generate textural explanations on academic networks, and the explanations are further leveraged to enhance the node features for conducting node classification. Besides enhancing the models with LLM-generated features, \citet{chen2023exploring} and \citet{ye2023natural} further propose to directly adopt LLMs for predicting node categories on text-attributed graphs. \citet{wang2023can} introduce a benchmark framework to evaluate the performance of LLMs with several graph algorithmic tasks, including topological sort, maximum flows, {\it etc}. \citet{guo2023gpt4graph} is another benchmark to evaluate LLMs' abilities to tackle structural information, and several factors, such as the role of graph description languages, are taken into consideration in the evaluation process. \citet{zhang2023LLM4DyG} further considers evaluating the LLMs' abilities to handle spatial-temporal information on dynamic graphs. Several other works explore the applications of LLMs in graph tool learning~\cite{zhang2023toolformer,jiang2023structgpt}, knowledge graphs~\cite{pan2023unifying}, {\it etc}. However, most existing works mainly focus on graph discriminative tasks, and we explore the potential of LLMs on graph generation, which remains under-explored in the literature.

\subsection{Graph generation}

Graph generation is a highly anticipated research field that is crucial for areas such as code generation and new drug discovery~\cite{zhu2022survey}. 
Recently, deep graph generation methods have achieved remarkable results, which can be roughly classified into auto-regressive models~\cite{GraphGen, Graphgen-redux}, variational autoencoder models~\cite{MDVAE,JT-VAE}, normalizing flow models~\cite{MoFlow,GraphDF}, generative adversarial network~\cite{MMGAN,Mol-CycleGAN}, diffusion models~\cite{GDSS,DiGress}. However, these methods mainly focus on learning the distribution of graphs from existing graphs for generation, ignoring the complex yet rich domain knowledge in practical applications. Inspired by the recent progress of LLMs in leveraging expert knowledge, in this paper, we systematically explore the graph generation capabilities of LLMs for the first time by proposing various tasks, including rule-based generation, distribution-based generation, and property-based generation.

\section{Conclusion}

In this paper, we explore the potential of large language models (LLMs) for graph generation tasks, which remains unexplored in the literature. We propose \model to systematically assess LLMs' capabilities in understanding and applying various graph structure rules, capturing structural type distributions, and leveraging domain knowledge in property-based graph generation. Our findings indicate that while LLMs show promising preliminary abilities in rule-based and distribution-based graph generation, the effectiveness of popular prompting methods like few-shot and chain-of-thought prompting is not consistent. Additionally, the potential of LLMs to generate molecules with specific properties is a notable outcome. These insights open new avenues for future research in graph generation, highlighting the evolving capabilities as well as the limitations of LLMs in graph generation tasks.

\bibliographystyle{unsrtnat}
\bibliography{references}

\begin{thebibliography}{41}
\providecommand{\natexlab}[1]{#1}
\providecommand{\url}[1]{\texttt{#1}}
\expandafter\ifx\csname urlstyle\endcsname\relax
  \providecommand{\doi}[1]{doi: #1}\else
  \providecommand{\doi}{doi: \begingroup \urlstyle{rm}\Url}\fi

\bibitem[Zhao et~al.(2023)Zhao, Zhou, Li, Tang, Wang, Hou, Min, Zhang, Zhang, Dong, et~al.]{zhao2023survey}
Wayne~Xin Zhao, Kun Zhou, Junyi Li, Tianyi Tang, Xiaolei Wang, Yupeng Hou, Yingqian Min, Beichen Zhang, Junjie Zhang, Zican Dong, et~al.
\newblock A survey of large language models.
\newblock \emph{arXiv preprint arXiv:2303.18223}, 2023.

\bibitem[Bommasani et~al.(2022)Bommasani, Hudson, Adeli, et~al.]{bommasani2022opportunities}
Rishi Bommasani, Drew~A. Hudson, Ehsan Adeli, et~al.
\newblock On the opportunities and risks of foundation models.
\newblock \emph{arXiv preprint arXiv:2303.18223}, 2022.

\bibitem[Brown et~al.(2020)Brown, Mann, Ryder, et~al.]{GPT3}
Tom Brown, Benjamin Mann, Nick Ryder, et~al.
\newblock Language models are few-shot learners.
\newblock In \emph{Advances in Neural Information Processing Systems}, pages 1877--1901, 2020.

\bibitem[Bubeck et~al.(2023)Bubeck, Chandrasekaran, Eldan, Gehrke, Horvitz, Kamar, Lee, Lee, Li, Lundberg, et~al.]{bubeck2023sparks}
S{\'e}bastien Bubeck, Varun Chandrasekaran, Ronen Eldan, Johannes Gehrke, Eric Horvitz, Ece Kamar, Peter Lee, Yin~Tat Lee, Yuanzhi Li, Scott Lundberg, et~al.
\newblock Sparks of artificial general intelligence: Early experiments with gpt-4.
\newblock \emph{arXiv preprint arXiv:2303.12712}, 2023.

\bibitem[Wei et~al.(2022)Wei, Wang, Schuurmans, Bosma, ichter, Xia, Chi, Le, and Zhou]{COT}
Jason Wei, Xuezhi Wang, Dale Schuurmans, Maarten Bosma, brian ichter, Fei Xia, Ed~Chi, Quoc~V Le, and Denny Zhou.
\newblock Chain-of-thought prompting elicits reasoning in large language models.
\newblock In S.~Koyejo, S.~Mohamed, A.~Agarwal, D.~Belgrave, K.~Cho, and A.~Oh, editors, \emph{Advances in Neural Information Processing Systems}, volume~35, pages 24824--24837, 2022.

\bibitem[Ni et~al.(2023)Ni, Iyer, Radev, Stoyanov, Yih, Wang, and Lin]{code1}
Ansong Ni, Srini Iyer, Dragomir Radev, Veselin Stoyanov, Wen-Tau Yih, Sida Wang, and Xi~Victoria Lin.
\newblock {LEVER}: Learning to verify language-to-code generation with execution.
\newblock In Andreas Krause, Emma Brunskill, Kyunghyun Cho, Barbara Engelhardt, Sivan Sabato, and Jonathan Scarlett, editors, \emph{Proceedings of the 40th International Conference on Machine Learning}, volume 202 of \emph{Proceedings of Machine Learning Research}, pages 26106--26128. PMLR, 2023.
\newblock URL \url{https://proceedings.mlr.press/v202/ni23b.html}.

\bibitem[Vaithilingam et~al.(2022)Vaithilingam, Zhang, and Glassman]{code2}
Priyan Vaithilingam, Tianyi Zhang, and Elena~L. Glassman.
\newblock Expectation vs. experience: Evaluating the usability of code generation tools powered by large language models.
\newblock In \emph{Extended Abstracts of the 2022 CHI Conference on Human Factors in Computing Systems}, CHI EA '22, New York, NY, USA, 2022. Association for Computing Machinery.
\newblock ISBN 9781450391566.
\newblock \doi{10.1145/3491101.3519665}.
\newblock URL \url{https://doi.org/10.1145/3491101.3519665}.

\bibitem[Murakumo et~al.(2023)Murakumo, Yoshikawa, Rikimaru, Nakamura, Furui, Suzuki, Yamasaki, Nishigaya, Takagi, and Ohue]{drug1}
Kusuri Murakumo, Naruki Yoshikawa, Kentaro Rikimaru, Shogo Nakamura, Kairi Furui, Takamasa Suzuki, Hiroyuki Yamasaki, Yuki Nishigaya, Yuzo Takagi, and Masahito Ohue.
\newblock {LLM} drug discovery challenge: A contest as a feasibility study on the utilization of large language models in medicinal chemistry.
\newblock In \emph{AI for Accelerated Materials Design - NeurIPS 2023 Workshop}, 2023.
\newblock URL \url{https://openreview.net/forum?id=kjUylvko18}.

\bibitem[Chakraborty et~al.(2023)Chakraborty, Bhattacharya, and Lee]{drug2}
Chiranjib Chakraborty, Manojit Bhattacharya, and Sang-Soo Lee.
\newblock Artificial intelligence enabled chatgpt and large language models in drug target discovery, drug discovery, and development.
\newblock \emph{Molecular Therapy-Nucleic Acids}, 33:\penalty0 866--868, 2023.

\bibitem[Meyer et~al.(2023)Meyer, Stadler, Frey, Radtke, Junghanns, Meissner, Dziwis, Bulert, and Martin]{know1}
Lars-Peter Meyer, Claus Stadler, Johannes Frey, Norman Radtke, Kurt Junghanns, Roy Meissner, Gordian Dziwis, Kirill Bulert, and Michael Martin.
\newblock Llm-assisted knowledge graph engineering: Experiments with chatgpt.
\newblock \emph{arXiv preprint arXiv:2307.06917}, 2023.

\bibitem[Sun et~al.(2023)Sun, Xu, Zha, Liu, and Dong]{know2}
Kai Sun, Yifan~Ethan Xu, Hanwen Zha, Yue Liu, and Xin~Luna Dong.
\newblock Head-to-tail: How knowledgeable are large language models (llm)? aka will llms replace knowledge graphs?
\newblock \emph{arXiv preprint arXiv:2308.10168}, 2023.

\bibitem[Zhang et~al.(2023{\natexlab{a}})Zhang, Li, Zhang, Qin, Wang, and Zhu]{zhang2023graph}
Ziwei Zhang, Haoyang Li, Zeyang Zhang, Yijian Qin, Xin Wang, and Wenwu Zhu.
\newblock Graph meets llms: Towards large graph models.
\newblock \emph{NeurIPS 2023 New Frontiers in Graph Learning Workshop}, 2023{\natexlab{a}}.

\bibitem[Wang et~al.(2023{\natexlab{a}})Wang, Feng, He, Tan, Han, and Tsvetkov]{wang2023can}
Heng Wang, Shangbin Feng, Tianxing He, Zhaoxuan Tan, Xiaochuang Han, and Yulia Tsvetkov.
\newblock Can language models solve graph problems in natural language?
\newblock \emph{Thirty-seventh Conference on Neural Information Processing Systems}, 2023{\natexlab{a}}.

\bibitem[Ye et~al.(2023)Ye, Zhang, Wang, Xu, and Zhang]{ye2023natural}
Ruosong Ye, Caiqi Zhang, Runhui Wang, Shuyuan Xu, and Yongfeng Zhang.
\newblock Natural language is all a graph needs.
\newblock \emph{arXiv preprint arXiv:2308.07134}, 2023.

\bibitem[Huang et~al.(2023)Huang, Zhang, Mei, and Ma]{huang2023llms}
Jin Huang, Xingjian Zhang, Qiaozhu Mei, and Jiaqi Ma.
\newblock Can llms effectively leverage graph structural information: When and why.
\newblock \emph{arXiv preprint arXiv:2309.16595}, 2023.

\bibitem[Tang et~al.(2023)Tang, Yang, Wei, Shi, Su, Cheng, Yin, and Huang]{tang2023graphgpt}
Jiabin Tang, Yuhao Yang, Wei Wei, Lei Shi, Lixin Su, Suqi Cheng, Dawei Yin, and Chao Huang.
\newblock Graphgpt: Graph instruction tuning for large language models.
\newblock \emph{arXiv preprint arXiv:2310.13023}, 2023.

\bibitem[Zhu et~al.(2022)Zhu, Du, Wang, Xu, Zhang, Liu, and Wu]{zhu2022survey}
Yanqiao Zhu, Yuanqi Du, Yinkai Wang, Yichen Xu, Jieyu Zhang, Qiang Liu, and Shu Wu.
\newblock A survey on deep graph generation: Methods and applications.
\newblock In \emph{Learning on Graphs Conference}, pages 47--1, 2022.

\bibitem[You et~al.(2018)You, Liu, Ying, Pande, and Leskovec]{you2018graph}
Jiaxuan You, Bowen Liu, Zhitao Ying, Vijay Pande, and Jure Leskovec.
\newblock Graph convolutional policy network for goal-directed molecular graph generation.
\newblock \emph{Advances in neural information processing systems}, 31, 2018.

\bibitem[Ying et~al.(2019)Ying, Bourgeois, You, Zitnik, and Leskovec]{gnnexplainer}
Zhitao Ying, Dylan Bourgeois, Jiaxuan You, Marinka Zitnik, and Jure Leskovec.
\newblock Gnnexplainer: Generating explanations for graph neural networks.
\newblock \emph{Advances in neural information processing systems}, 32, 2019.

\bibitem[Wu et~al.(2022)Wu, Wang, Zhang, He, and Chua]{dir}
Yingxin Wu, Xiang Wang, An~Zhang, Xiangnan He, and Tat-Seng Chua.
\newblock Discovering invariant rationales for graph neural networks.
\newblock In \emph{International Conference on Learning Representations}, 2022.

\bibitem[Qin et~al.(2022)Qin, Wang, Zhang, Xie, and Zhu]{Grace}
Yijian Qin, Xin Wang, Ziwei Zhang, Pengtao Xie, and Wenwu Zhu.
\newblock Graph neural architecture search under distribution shifts.
\newblock In Kamalika Chaudhuri, Stefanie Jegelka, Le~Song, Csaba Szepesv{\'{a}}ri, Gang Niu, and Sivan Sabato, editors, \emph{International Conference on Machine Learning, {ICML} 2022, 17-23 July 2022, Baltimore, Maryland, {USA}}, volume 162 of \emph{Proceedings of Machine Learning Research}, pages 18083--18095. {PMLR}, 2022.
\newblock URL \url{https://proceedings.mlr.press/v162/qin22b.html}.

\bibitem[Guo and Zhao(2023)]{ggsurvey}
Xiaojie Guo and Liang Zhao.
\newblock A systematic survey on deep generative models for graph generation.
\newblock \emph{{IEEE} Trans. Pattern Anal. Mach. Intell.}, 45\penalty0 (5):\penalty0 5370--5390, 2023.
\newblock \doi{10.1109/TPAMI.2022.3214832}.
\newblock URL \url{https://doi.org/10.1109/TPAMI.2022.3214832}.

\bibitem[Hu et~al.(2020)Hu, Fey, Zitnik, Dong, Ren, Liu, Catasta, and Leskovec]{hu2020open}
Weihua Hu, Matthias Fey, Marinka Zitnik, Yuxiao Dong, Hongyu Ren, Bowen Liu, Michele Catasta, and Jure Leskovec.
\newblock Open graph benchmark: Datasets for machine learning on graphs.
\newblock \emph{Advances in neural information processing systems}, 33:\penalty0 22118--22133, 2020.

\bibitem[Weininger(1988)]{weininger1988smiles}
David Weininger.
\newblock Smiles, a chemical language and information system. 1. introduction to methodology and encoding rules.
\newblock \emph{Journal of chemical information and computer sciences}, 28\penalty0 (1):\penalty0 31--36, 1988.

\bibitem[He et~al.(2023)He, Bresson, Laurent, and Hooi]{he2023explanations}
Xiaoxin He, Xavier Bresson, Thomas Laurent, and Bryan Hooi.
\newblock Explanations as features: Llm-based features for text-attributed graphs.
\newblock \emph{arXiv preprint arXiv:2305.19523}, 2023.

\bibitem[Chen et~al.(2023)Chen, Mao, Li, Jin, Wen, Wei, Wang, Yin, Fan, Liu, and Tang]{chen2023exploring}
Zhikai Chen, Haitao Mao, Hang Li, Wei Jin, Hongzhi Wen, Xiaochi Wei, Shuaiqiang Wang, Dawei Yin, Wenqi Fan, Hui Liu, and Jiliang Tang.
\newblock Exploring the potential of large language models (llms) in learning on graphs.
\newblock \emph{arXiv preprint arXiv:2307.03393}, 2023.

\bibitem[Guo et~al.(2023)Guo, Du, and Liu]{guo2023gpt4graph}
Jiayan Guo, Lun Du, and Hengyu Liu.
\newblock Gpt4graph: Can large language models understand graph structured data? an empirical evaluation and benchmarking.
\newblock \emph{arXiv preprint arXiv:2305.15066}, 2023.

\bibitem[Zhang et~al.(2023{\natexlab{b}})Zhang, Wang, Zhang, Li, Qin, Wu, and Zhu]{zhang2023LLM4DyG}
Zeyang Zhang, Xin Wang, Ziwei Zhang, Haoyang Li, Yijian Qin, Simin Wu, and Wenwu Zhu.
\newblock Llm4dyg: Can large language models solve problems on dynamic graphs?
\newblock \emph{arXiv preprint arXiv:2310.17110}, 2023{\natexlab{b}}.

\bibitem[Zhang(2023)]{zhang2023toolformer}
Jiawei Zhang.
\newblock Graph-toolformer: To empower llms with graph reasoning ability via prompt augmented by chatgpt.
\newblock \emph{arXiv preprint arXiv:2304.11116}, 2023.

\bibitem[Jiang et~al.(2023)Jiang, Zhou, Dong, Ye, Zhao, and Wen]{jiang2023structgpt}
Jinhao Jiang, Kun Zhou, Zican Dong, Keming Ye, Wayne~Xin Zhao, and Ji-Rong Wen.
\newblock Structgpt: A general framework for large language model to reason over structured data.
\newblock \emph{arXiv preprint arXiv:2305.09645}, 2023.

\bibitem[Pan et~al.(2023)Pan, Luo, Wang, Chen, Wang, and Wu]{pan2023unifying}
Shirui Pan, Linhao Luo, Yufei Wang, Chen Chen, Jiapu Wang, and Xindong Wu.
\newblock Unifying large language models and knowledge graphs: A roadmap.
\newblock \emph{arXiv preprint arXiv:2306.08302}, 2023.

\bibitem[Wang et~al.(2023{\natexlab{b}})Wang, Feng, He, Tan, Han, and Tsvetkov]{GraphGen}
Heng Wang, Shangbin Feng, Tianxing He, Zhaoxuan Tan, Xiaochuang Han, and Yulia Tsvetkov.
\newblock Can language models solve graph problems in natural language?
\newblock \emph{CoRR}, abs/2305.10037, 2023{\natexlab{b}}.
\newblock \doi{10.48550/arXiv.2305.10037}.
\newblock URL \url{https://doi.org/10.48550/arXiv.2305.10037}.

\bibitem[Bacciu and Podda(2021)]{Graphgen-redux}
Davide Bacciu and Marco Podda.
\newblock Graphgen-redux: a fast and lightweight recurrent model for labeled graph generation.
\newblock In \emph{International Joint Conference on Neural Networks, {IJCNN} 2021, Shenzhen, China, July 18-22, 2021}, pages 1--8. {IEEE}, 2021.
\newblock \doi{10.1109/IJCNN52387.2021.9533743}.
\newblock URL \url{https://doi.org/10.1109/IJCNN52387.2021.9533743}.

\bibitem[Du et~al.()Du, Guo, Shehu, and Zhao]{MDVAE}
Yuanqi Du, Xiaojie Guo, Amarda Shehu, and Liang Zhao.
\newblock \emph{Interpretable Molecular Graph Generation via Monotonic Constraints}, pages 73--81.
\newblock \doi{10.1137/1.9781611977172.9}.
\newblock URL \url{https://epubs.siam.org/doi/abs/10.1137/1.9781611977172.9}.

\bibitem[Jin et~al.(2018)Jin, Barzilay, and Jaakkola]{JT-VAE}
Wengong Jin, Regina Barzilay, and Tommi~S. Jaakkola.
\newblock Junction tree variational autoencoder for molecular graph generation.
\newblock In Jennifer~G. Dy and Andreas Krause, editors, \emph{Proceedings of the 35th International Conference on Machine Learning, {ICML} 2018, Stockholmsm{\"{a}}ssan, Stockholm, Sweden, July 10-15, 2018}, volume~80 of \emph{Proceedings of Machine Learning Research}, pages 2328--2337. {PMLR}, 2018.
\newblock URL \url{http://proceedings.mlr.press/v80/jin18a.html}.

\bibitem[Zang and Wang(2020)]{MoFlow}
Chengxi Zang and Fei Wang.
\newblock Moflow: An invertible flow model for generating molecular graphs.
\newblock In Rajesh Gupta, Yan Liu, Jiliang Tang, and B.~Aditya Prakash, editors, \emph{{KDD} '20: The 26th {ACM} {SIGKDD} Conference on Knowledge Discovery and Data Mining, Virtual Event, CA, USA, August 23-27, 2020}, pages 617--626. {ACM}, 2020.
\newblock \doi{10.1145/3394486.3403104}.
\newblock URL \url{https://doi.org/10.1145/3394486.3403104}.

\bibitem[Luo et~al.(2021)Luo, Yan, and Ji]{GraphDF}
Youzhi Luo, Keqiang Yan, and Shuiwang Ji.
\newblock Graphdf: {A} discrete flow model for molecular graph generation.
\newblock In Marina Meila and Tong Zhang, editors, \emph{Proceedings of the 38th International Conference on Machine Learning, {ICML} 2021, 18-24 July 2021, Virtual Event}, volume 139 of \emph{Proceedings of Machine Learning Research}, pages 7192--7203. {PMLR}, 2021.
\newblock URL \url{http://proceedings.mlr.press/v139/luo21a.html}.

\bibitem[Gamage et~al.(2020)Gamage, Chien, Peng, and Milenkovic]{MMGAN}
Anuththari Gamage, Eli Chien, Jianhao Peng, and Olgica Milenkovic.
\newblock Multi-motifgan {(MMGAN):} motif-targeted graph generation and prediction.
\newblock In \emph{2020 {IEEE} International Conference on Acoustics, Speech and Signal Processing, {ICASSP} 2020, Barcelona, Spain, May 4-8, 2020}, pages 4182--4186. {IEEE}, 2020.
\newblock \doi{10.1109/ICASSP40776.2020.9053451}.
\newblock URL \url{https://doi.org/10.1109/ICASSP40776.2020.9053451}.

\bibitem[Maziarka et~al.(2020)Maziarka, Pocha, Kaczmarczyk, Rataj, Danel, and Warchol]{Mol-CycleGAN}
Lukasz Maziarka, Agnieszka Pocha, Jan Kaczmarczyk, Krzysztof Rataj, Tomasz Danel, and Michal Warchol.
\newblock Mol-cyclegan: a generative model for molecular optimization.
\newblock \emph{J. Cheminformatics}, 12\penalty0 (1):\penalty0 2, 2020.
\newblock \doi{10.1186/S13321-019-0404-1}.
\newblock URL \url{https://doi.org/10.1186/s13321-019-0404-1}.

\bibitem[Jo et~al.(2022)Jo, Lee, and Hwang]{GDSS}
Jaehyeong Jo, Seul Lee, and Sung~Ju Hwang.
\newblock Score-based generative modeling of graphs via the system of stochastic differential equations.
\newblock In Kamalika Chaudhuri, Stefanie Jegelka, Le~Song, Csaba Szepesv{\'{a}}ri, Gang Niu, and Sivan Sabato, editors, \emph{International Conference on Machine Learning, {ICML} 2022, 17-23 July 2022, Baltimore, Maryland, {USA}}, volume 162 of \emph{Proceedings of Machine Learning Research}, pages 10362--10383. {PMLR}, 2022.
\newblock URL \url{https://proceedings.mlr.press/v162/jo22a.html}.

\bibitem[Vignac et~al.(2023)Vignac, Krawczuk, Siraudin, Wang, Cevher, and Frossard]{DiGress}
Cl{\'{e}}ment Vignac, Igor Krawczuk, Antoine Siraudin, Bohan Wang, Volkan Cevher, and Pascal Frossard.
\newblock Digress: Discrete denoising diffusion for graph generation.
\newblock In \emph{The Eleventh International Conference on Learning Representations, {ICLR} 2023, Kigali, Rwanda, May 1-5, 2023}. OpenReview.net, 2023.
\newblock URL \url{https://openreview.net/pdf?id=UaAD-Nu86WX}.

\end{thebibliography}

\clearpage
\appendix

\section{Experimental setup}
\subsection{Rule-based generation}
For the rule-based graph generation experiments, we configure the tasks with the problem settings listed in Table \ref{tab:rule-settings} unless otherwise specified. The task settings are chosen to be challenging for LLMs. Additionally, to produce the results in the comparison of different graph sizes, the parameters for the three different sizes are listed in Table \ref{tab:rule-settings-size}; note that the ``Medium'' size is the same as the default settings. In the comparison of different LLMs, since models other than GPT-4 are relatively weak, we use the ``Small'' settings from Table \ref{tab:rule-settings-size} in the experiments. Unless otherwise specified, we use the sampling temperature $t=0.8$.

It is worth noting that the number of edges have a significant effect on the difficulty of ``Planar'' and ``$k$-color'' tasks, as graphs with few edges are easily planar or $k$-colorable. The parameters in the experiments are chosen such that random graphs with the specified number of nodes and edges have around 20\% probability to be valid.

\begin{table}[h]
  \caption{Problem settings for different rule-based tasks.}
  \label{tab:rule-settings}
  \centering
  \begin{tabular}{cc}
    \toprule
    Task & Settings \\
    \midrule
    Trees & 15 nodes \\
    Cycles & 15 nodes \\
    Components & 15 nodes, 5 components \\
    Planar & 15 nodes, 24 edges \\
    $k$-regular & 16 nodes, $k=3$ \\
    Wheel & 15 nodes \\
    Bipartite & 5 nodes in each partition \\
    $k$-color & 15 nodes, 32 edges, $k=3$ \\
    \bottomrule
  \end{tabular}
\end{table}

\begin{table}[h]
  \caption{Problem settings for rule-based tasks with different sizes.}
  \label{tab:rule-settings-size}
  \centering
  \begin{tabular}{ccc}
    \toprule
    \multicolumn{2}{c}{Task} & Settings \\
    \midrule
    \multirow{3}{*}{Cycles} & Small & 10 nodes \\
    & Medium & 15 nodes \\
    & Large & 20 nodes \\
    \midrule
    \multirow{3}{*}{$k$-regular} & Small & 12 nodes, $k=3$ \\
    & Medium & 16 nodes, $k=3$ \\
    & Large & 20 nodes, $k=3$ \\
    \midrule
    \multirow{3}{*}{$k$-color} & Small & 10 nodes, 20 edges, $k=3$ \\
    & Medium & 15 nodes, 32 edges, $k=3$ \\
    & Large & 18 nodes, 39 edges, $k=3$ \\
    \bottomrule
  \end{tabular}
\end{table}

\subsection{Distribution-based generation}
For the ``Trees or cycles'' task, the number of nodes for each tree or cycle is randomly selected between 5 and 7. For the ``Union of components'' task, each of the two components for each graph has between 5 and 7 nodes. For the ``Motif'' task, we define the three kinds of base graphs and motif graphs in the same way as Spurious-Motif, and the description is given in Table \ref{tab:motif-def}. All experiments use the sampling temperature $t=0.5$.

\begin{table}
    \caption{The definition of base graphs and motif graphs for the ``Motif'' task. The definition of motif graphs are given by first listing the number of nodes, then listing the endpoints of all edges, which is in the same format as the input to LLMs.}
    \label{tab:motif-def}
    \centering
    \begin{tabular}{cl}
        \toprule
        Graph & Description \\
        \midrule
        Trees & Full binary or ternary tree \\
        Ladders & Two paths of same length with edge connections between pairs of nodes \\
        Wheels & A single node connected to all nodes of a cycle \\
        \midrule
        Cycle & (5, [(1, 2), (1, 5), (2, 3), (3, 4), (4, 5)]) \\
        House & (5, [(1, 2), (1, 5), (2, 3), (2, 5), (3, 4), (4, 5)]) \\
        Crane & (5, [(1, 2), (1, 3), (1, 4), (1, 5), (2, 3), (3, 4), (4, 5)]) \\
        \bottomrule
    \end{tabular}
\end{table}

\subsection{Property-based generation}

\begin{table}
    \caption{The confusion matrix of the molecule classifier used in the experiments of property-based graph generation. $C(G)$ denotes the ground-truth label, $C_M(G)$ denotes the predicted label, and the numbers represent the amount of molecules that fall into this category.}
    \label{tab:molhiv-classifier}
    \centering
    \begin{tabular}{ccc}
        \toprule
        & $C_M(G)=0$ & $C_M(G)=1$ \\
        \midrule
        $C(G)=0$ & 35539 & 4145 \\
        $C(G)=1$ & 633 & 810 \\
        \bottomrule
    \end{tabular}
\end{table}

For property-based graph generation, we generate molecules using GPT-4 with sampling temperature set to $t=0.5$. To evaluate the performance of molecule generation, we measure the fraction of generated molecules that have the desired property using a GNN classifier trained on the OGBG-MolHIV dataset.
More specifically, let $G_1, G_2, \ldots, G_n$ be the generate graphs that represent molecules. Let $C_T(G) \in \{0, 1\}$ be the ground-truth label of graph $G$, with $C_T(G) = 1$ meaning the molecule $G$ has the desired property (e.g. can inhibit HIV replication), and $C_T(G) = 0$ meaning otherwise. Similarly, let $C_M(G)$ be the predicted label of $G$. These values are related using the law of total probability:
\begin{small}
\begin{equation*}
    p(C_M(G)=1) = p(C_T(G)=0) p(C_M(G)=1 \mid C_T(G)=0) + p(C_T(G)=1) p(C_M(G)=1 \mid C_T(G)=1).
\end{equation*}
\end{small}
where 
the conditional probability can be computed from the confusion matrix of the classifier as given in Table \ref{tab:molhiv-classifier}. By rearranging the equations, we can compute rectified predictions $C(G)$ that are closer to $C_T(G)$:
\begin{small}
\begin{equation*}
    p(C(G)=1) =
    \frac{p(C_M(G)=1) - p(C_M(G)=1 \mid C_T(G)=0) }{p(C_M(G)=1 \mid C_T(G)=1) - p(C_M(G)=1 \mid C_T(G)=0)}.
\end{equation*}
\end{small}

\section{Extended experimental results}
We provide complete experimental results that include valid rates, novel rates, and unique rates for rule-based generation tasks in Table \ref{tab:rule-total-appendix}, Table \ref{tab:rule-size-appendix}, and Table \ref{tab:rule-llm-appendix}. Additionally, we compare the effect of different sampling temperatures for rule-based generation in Table \ref{tab:rule-temperature}, and the effect of different amounts of generated graphs in Table \ref{tab:rule-ansnum}. From the two tables, we find that it is necessary to set appropriate sampling temperature and amount of generated graphs for different tasks.

For distribution-based generation, we provide numerical results in Table \ref{tab:dist-task1}, Table \ref{tab:dist-task2} and Table \ref{tab:dist-task3}.

\begin{table}[b]
  \caption{Metrics for distribution-based graph generation on the ``Trees or cycles'' task. $p$ represents the parameter of the distribution where the graphs are sampled. $p_{\text{pred}}$ is the value of $p$ predicted by LLM for the input graphs. $p_{\text{gen}}$ is the value of $p$ calculated by the generated graphs. ``Valid'' denotes fractions of generated graphs that are valid under specific rules.}
  \label{tab:dist-task1}
  \centering
  \adjustbox{max width=\linewidth}{
  \begin{tabular}{ccccc}
    \toprule
    $p$&Prompt&$p_{\text{pred}}$&$p_{\text{gen}}$&Valid\\
    \midrule
    \multirow{3}{*}{$p=0.2$} & Zero-shot & $62.0 \pm 1.8$ & $54.3 \pm 17.8$ & $72.0 \pm 7.2$\\
 & Few-shot & $60.0 \pm 0.0$ & $32.2 \pm 14.9$ & $86.0 \pm 6.1$\\
 & CoT & $45.0 \pm 3.5$ & $50.4 \pm 6.7$ & $83.3 \pm 9.8$\\
\midrule
\multirow{3}{*}{$p=0.4$} & Zero-shot & $60.0 \pm 2.8$ & $74.3 \pm 12.7$ & $84.0 \pm 5.4$\\
 & Few-shot & $62.0 \pm 1.8$ & $68.9 \pm 16.1$ & $98.0 \pm 1.8$\\
 & CoT & $43.3 \pm 2.7$ & $59.0 \pm 4.8$ & $83.3 \pm 7.2$\\
\midrule
\multirow{3}{*}{$p=0.6$} & Zero-shot & $58.0 \pm 1.8$ & $48.1 \pm 15.1$ & $88.0 \pm 5.2$\\
 & Few-shot & $62.0 \pm 1.8$ & $16.0 \pm 10.4$ & $92.0 \pm 5.2$\\
 & CoT & $53.3 \pm 5.4$ & $86.3 \pm 7.1$ & $96.7 \pm 2.7$\\
\midrule
\multirow{3}{*}{$p=0.8$} & Zero-shot & $60.0 \pm 0.0$ & $90.0 \pm 8.9$ & $92.0 \pm 5.2$\\
 & Few-shot & $64.0 \pm 2.2$ & $62.1 \pm 12.1$ & $88.0 \pm 5.2$\\
 & CoT & $76.7 \pm 5.4$ & $92.6 \pm 6.0$ & $93.3 \pm 2.7$\\
    \bottomrule
  \end{tabular}
  }
\end{table}

\begin{table}
  \caption{Metrics for distribution-based graph generation on the ``Union of components'' task. $p$ represents the parameter of the distribution where the graphs are sampled. $p_{\text{pred}}$ is the value of $p$ predicted by LLM for the input graphs. $p_{\text{gen}}$ is the value of $p$ calculated by the generated graphs. ``Valid'' denotes fractions of generated graphs that are valid under specific rules.}
  \label{tab:dist-task2}
  \centering
  \adjustbox{max width=\linewidth}{
  \begin{tabular}{ccccc}
    \toprule
    $p$&Prompt&$p_{\text{pred}}$&$p_{\text{gen}}$&Valid\\
    \midrule
    \multirow{3}{*}{$p=0.2$} & Zero-shot & $62.0 \pm 1.8$ & $58.2 \pm 16.2$ & $82.0 \pm 4.4$\\
 & Few-shot & $60.0 \pm 0.0$ & $40.3 \pm 14.9$ & $70.0 \pm 11.7$\\
 & CoT & $30.0 \pm 8.5$ & $25.0 \pm 10.0$ & $46.0 \pm 8.8$\\
\midrule
\multirow{3}{*}{$p=0.4$} & Zero-shot & $58.0 \pm 1.8$ & $57.2 \pm 16.3$ & $70.0 \pm 13.3$\\
 & Few-shot & $60.0 \pm 0.0$ & $62.2 \pm 12.8$ & $84.0 \pm 6.1$\\
 & CoT & $36.0 \pm 3.6$ & $42.0 \pm 7.9$ & $68.0 \pm 4.4$\\
\midrule
\multirow{3}{*}{$p=0.6$} & Zero-shot & $62.0 \pm 1.8$ & $49.2 \pm 16.1$ & $72.0 \pm 14.8$\\
 & Few-shot & $60.0 \pm 0.0$ & $52.8 \pm 14.8$ & $86.0 \pm 4.6$\\
 & CoT & $53.4 \pm 7.0$ & $70.6 \pm 8.6$ & $72.0 \pm 9.5$\\
\midrule
\multirow{3}{*}{$p=0.8$} & Zero-shot & $58.0 \pm 1.8$ & $72.0 \pm 11.1$ & $54.0 \pm 16.9$\\
 & Few-shot & $60.0 \pm 0.0$ & $45.0 \pm 15.2$ & $86.0 \pm 3.6$\\
 & CoT & $59.2 \pm 5.0$ & $62.9 \pm 9.7$ & $78.0 \pm 5.2$\\
    \bottomrule
  \end{tabular}
  }
\end{table}

\begin{table}
  \caption{Metrics for distribution-based graph generation on the ``Motif'' task. $p$ represents the parameter of the distribution where the graphs are sampled. $p_{\text{pred}}$ is the value of $p$ predicted by LLM for the input graphs. $p_{\text{gen}}$ is the value of $p$ calculated by the generated graphs. ``Valid'' denotes fractions of generated graphs that are valid under specific rules.}
  \label{tab:dist-task3}
  \centering
  \adjustbox{max width=\linewidth}{
  \begin{tabular}{ccccc}
    \toprule
    $p$&Prompt&$p_{\text{pred}}$&$p_{\text{gen}}$&Valid\\
    \midrule
    \multirow{3}{*}{$p=0.2$} & Zero-shot & $60.0 \pm 0.0$ & $18.7 \pm 7.6$ & $56.0 \pm 18.9$\\
 & Few-shot & $48.0 \pm 5.0$ & $42.0 \pm 14.8$ & $100.0 \pm 0.0$\\
 & CoT & $42.0 \pm 3.3$ & $42.1 \pm 2.4$ & $90.0 \pm 4.9$\\
\midrule
\multirow{3}{*}{$p=0.4$} & Zero-shot & $58.0 \pm 1.8$ & $22.9 \pm 5.9$ & $98.0 \pm 1.8$\\
 & Few-shot & $42.0 \pm 4.4$ & $33.7 \pm 10.5$ & $94.0 \pm 5.4$\\
 & CoT & $30.0 \pm 2.8$ & $55.6 \pm 9.0$ & $98.0 \pm 1.8$\\
\midrule
\multirow{3}{*}{$p=0.6$} & Zero-shot & $52.0 \pm 1.8$ & $24.2 \pm 7.3$ & $90.0 \pm 6.9$\\
 & Few-shot & $48.0 \pm 8.2$ & $38.9 \pm 10.0$ & $98.0 \pm 1.8$\\
 & CoT & $40.0 \pm 2.4$ & $52.2 \pm 6.1$ & $95.0 \pm 3.1$\\
\midrule
\multirow{3}{*}{$p=0.8$} & Zero-shot & $56.0 \pm 2.2$ & $35.1 \pm 11.2$ & $98.0 \pm 1.8$\\
 & Few-shot & $56.0 \pm 6.7$ & $38.0 \pm 8.2$ & $100.0 \pm 0.0$\\
 & CoT & $41.4 \pm 3.7$ & $63.3 \pm 7.7$ & $94.0 \pm 3.6$\\
    \bottomrule
  \end{tabular}
  }
\end{table}

\begin{table*}
    \centering
  \caption{The novel rate for rule-based graph generation with GPT-4. The metric measures the fraction of generated graphs that are different from the given example graphs. Values after $\pm$ denote standard errors.}
  \label{tab:rule-total-appendix}
  \adjustbox{max width=\linewidth}{
  \begin{tabular}{ccccccccccc}
    \toprule
    Metric & Prompt&Trees&Cycles& Components &Planar &$k$-regular&Wheel&Bipartite&$k$-color\\
    \midrule
    \multirow{4}{*}{Valid} & Zero-shot & $100.0 \pm 0.0$ & $91.3 \pm 3.3$ & $30.4 \pm 5.1$ & $47.3 \pm 4.2$ & $64.0 \pm 6.8$ & $13.0 \pm 5.3$ & $60.3 \pm 7.4$ & $50.3 \pm 5.5$ \\
    & Few-shot & $98.0 \pm 0.9$ & $85.0 \pm 3.3$ & $63.2 \pm 5.3$ & $4.3 \pm 1.3$ & $86.1 \pm 3.1$ & $88.8 \pm 7.4$ & $57.1 \pm 8.6$ & $62.3 \pm 5.1$ \\
    & Zero-shot+CoT & $100.0 \pm 0.0$ & $86.9 \pm 3.6$ & $38.0 \pm 5.1$ & $53.3 \pm 6.0$ & $82.7 \pm 8.6$ & $92.3 \pm 4.7$ & $92.7 \pm 4.4$ & $43.2 \pm 4.9$ \\
    & Few-shot+CoT & $97.6 \pm 1.7$ & $97.0 \pm 1.9$ & $40.0 \pm 6.7$ & $20.0 \pm 4.3$ & $91.5 \pm 1.6$ & $90.7 \pm 5.1$ & $98.2 \pm 1.8$ & $58.5 \pm 5.9$ \\
    \midrule
    \multirow{4}{*}{Unique} & Zero-shot  & $98.6 \pm 0.9$ & $88.3 \pm 1.6$ & $91.9 \pm 3.0$ & $99.7 \pm 0.3$ & $100.0 \pm 0.0$ & $98.7 \pm 0.8$ & $44.3 \pm 7.0$ & $98.3 \pm 1.6$ \\
    & Few-shot  & $99.3 \pm 0.5$ & $92.3 \pm 1.3$ & $97.7 \pm 1.1$ & $96.8 \pm 2.5$ & $100.0 \pm 0.0$ & $98.8 \pm 1.1$ & $50.0 \pm 8.2$ & $98.5 \pm 1.5$ \\
    & Zero-shot+CoT  & $100.0 \pm 0.0$ & $89.0 \pm 4.2$ & $98.5 \pm 0.8$ & $98.6 \pm 0.8$ & $100.0 \pm 0.0$ & $85.0 \pm 5.7$ & $16.9 \pm 4.2$ & $98.6 \pm 1.0$ \\
    & Few-shot+CoT & $99.7 \pm 0.3$ & $83.0 \pm 5.1$ & $96.3 \pm 1.5$ & $98.0 \pm 1.4$ & $100.0 \pm 0.0$ & $88.3 \pm 5.1$ & $10.7 \pm 0.7$ & $96.3 \pm 2.6$ \\
    \midrule
    \multirow{4}{*}{Novel} & Zero-shot & $100.0 \pm 0.0$ & $100.0 \pm 0.0$ & $100.0 \pm 0.0$ & $100.0 \pm 0.0$ & $100.0 \pm 0.0$ & $100.0 \pm 0.0$ & $100.0 \pm 0.0$ & $100.0 \pm 0.0$ \\
    & Few-shot & $100.0 \pm 0.0$ & $100.0 \pm 0.0$ & $100.0 \pm 0.0$ & $100.0 \pm 0.0$ & $100.0 \pm 0.0$ & $100.0 \pm 0.0$ & $95.8 \pm 4.1$ & $100.0 \pm 0.0$ \\
    & Zero-shot+CoT & $100.0 \pm 0.0$ & $100.0 \pm 0.0$ & $100.0 \pm 0.0$ & $100.0 \pm 0.0$ & $100.0 \pm 0.0$ & $100.0 \pm 0.0$ & $100.0 \pm 0.0$ & $100.0 \pm 0.0$ \\
    & Few-shot+CoT & $100.0 \pm 0.0$ & $100.0 \pm 0.0$ & $100.0 \pm 0.0$ & $100.0 \pm 0.0$ & $100.0 \pm 0.0$ & $100.0 \pm 0.0$ & $96.4 \pm 3.5$ & $100.0 \pm 0.0$ \\
    \bottomrule
  \end{tabular}
  }
\end{table*}

\begin{table*}
  \caption{Metrics about the comparison of different graph sizes for rule-based graph generation with GPT-4. ``Valid'' denotes fractions of generated graphs that are valid under specific rules, ``Unique'' denotes fractions of generated graphs that are not duplicates, and ``Novel`` denotes fractions of generated graphs that are different from the given example graphs.}
  \label{tab:rule-size-appendix}
  \centering
  \adjustbox{max width=\linewidth}{
  \begin{tabular}{cccccccccccc}
    \toprule
    \multirow{2}{*}{Size} &\multirow{2}{*}{Prompt}& \multicolumn{3}{c}{Cycles}& \multicolumn{3}{c}{$k$-regular}& \multicolumn{3}{c}{$k$-color}\\
    &&Valid&Unique&Novel&Valid&Unique&Novel&Valid&Unique&Novel\\
    \midrule
    \multirow{4}{*}{Small} & Zero-shot & $84.7 \pm 5.9$ & $90.0 \pm 3.1$ & $100.0 \pm 0.0$ & $74.7 \pm 9.9$ & $100.0 \pm 0.0$ & $100.0 \pm 0.0$ & $56.0 \pm 4.5$ & $100.0 \pm 0.0$ & $100.0 \pm 0.0$ \\
     & Few-shot & $73.3 \pm 7.1$ & $96.0 \pm 1.8$ & $100.0 \pm 0.0$ & $90.0 \pm 2.9$ & $100.0 \pm 0.0$ & $100.0 \pm 0.0$ & $76.4 \pm 5.3$ & $100.0 \pm 0.0$ & $100.0 \pm 0.0$ \\
     & Zero-shot+CoT & $95.3 \pm 3.2$ & $84.7 \pm 4.8$ & $100.0 \pm 0.0$ & $97.1 \pm 1.2$ & $100.0 \pm 0.0$ & $100.0 \pm 0.0$ & $62.0 \pm 6.1$ & $96.7 \pm 3.2$ & $100.0 \pm 0.0$ \\
     & Few-shot+CoT & $96.0 \pm 2.8$ & $94.7 \pm 2.1$ & $100.0 \pm 0.0$ & $87.9 \pm 3.4$ & $100.0 \pm 0.0$ & $100.0 \pm 0.0$ & $80.7 \pm 3.2$ & $100.0 \pm 0.0$ & $100.0 \pm 0.0$ \\
    \midrule
    \multirow{4}{*}{Medium} & Zero-shot & $91.3 \pm 3.3$ & $88.3 \pm 1.6$ & $100.0 \pm 0.0$ & $64.0 \pm 6.8$ & $100.0 \pm 0.0$ & $100.0 \pm 0.0$ & $50.3 \pm 5.5$ & $98.3 \pm 1.6$ & $100.0 \pm 0.0$ \\
     & Few-shot & $85.0 \pm 3.3$ & $92.3 \pm 1.3$ & $100.0 \pm 0.0$ & $86.1 \pm 3.1$ & $100.0 \pm 0.0$ & $100.0 \pm 0.0$ & $62.3 \pm 5.1$ & $98.5 \pm 1.5$ & $100.0 \pm 0.0$ \\
     & Zero-shot+CoT & $86.9 \pm 3.6$ & $89.0 \pm 4.2$ & $100.0 \pm 0.0$ & $82.7 \pm 8.6$ & $100.0 \pm 0.0$ & $100.0 \pm 0.0$ & $43.2 \pm 4.9$ & $98.6 \pm 1.0$ & $100.0 \pm 0.0$ \\
     & Few-shot+CoT & $97.0 \pm 1.9$ & $83.0 \pm 5.1$ & $100.0 \pm 0.0$ & $91.5 \pm 1.6$ & $100.0 \pm 0.0$ & $100.0 \pm 0.0$ & $58.5 \pm 5.9$ & $96.3 \pm 2.6$ & $100.0 \pm 0.0$ \\
    \midrule
    \multirow{4}{*}{Large} & Zero-shot & $96.0 \pm 1.6$ & $84.0 \pm 4.4$ & $100.0 \pm 0.0$ & $61.3 \pm 10.6$ & $100.0 \pm 0.0$ & $100.0 \pm 0.0$ & $44.3 \pm 6.8$ & $97.9 \pm 1.5$ & $100.0 \pm 0.0$ \\
     & Few-shot & $82.0 \pm 6.7$ & $96.7 \pm 2.0$ & $100.0 \pm 0.0$ & $70.0 \pm 7.4$ & $100.0 \pm 0.0$ & $100.0 \pm 0.0$ & $64.0 \pm 8.8$ & $74.0 \pm 13.7$ & $100.0 \pm 0.0$ \\
     & Zero-shot+CoT & $100.0 \pm 0.0$ & $85.6 \pm 9.0$ & $100.0 \pm 0.0$ & $70.0 \pm 21.2$ & $100.0 \pm 0.0$ & $100.0 \pm 0.0$ & $58.9 \pm 6.4$ & $95.6 \pm 4.2$ & $100.0 \pm 0.0$ \\
     & Few-shot+CoT & $95.0 \pm 3.6$ & $85.7 \pm 6.2$ & $100.0 \pm 0.0$ & $82.5 \pm 7.1$ & $100.0 \pm 0.0$ & $100.0 \pm 0.0$ & $53.3 \pm 10.8$ & $89.2 \pm 5.5$ & $100.0 \pm 0.0$ \\

    \bottomrule
  \end{tabular}
  }
\end{table*}

\begin{table*}
  \caption{Metrics about the comparison of rule-based graph generation using different LLMs. ``Valid'' denotes fractions of generated graphs that are valid under specific rules, ``Unique'' denotes fractions of generated graphs that are not duplicates, and ``Novel`` denotes fractions of generated graphs that are different from the given example graphs.}
  \label{tab:rule-llm-appendix}
  \centering
  \adjustbox{max width=\linewidth}{
  \begin{tabular}{cccccccccccc}
    \toprule
    \multirow{2}{*}{Model} &\multirow{2}{*}{Prompt}& \multicolumn{3}{c}{Cycles}& \multicolumn{3}{c}{$k$-regular}& \multicolumn{3}{c}{$k$-color}\\
    &&Valid&Unique&Novel&Valid&Unique&Novel&Valid&Unique&Novel\\
    \midrule

    \multirow{4}{*}{GPT-4} & Zero-shot & $84.7 \pm 5.9$ & $90.0 \pm 3.1$ & $100.0 \pm 0.0$ & $74.7 \pm 9.9$ & $100.0 \pm 0.0$ & $100.0 \pm 0.0$ & $56.0 \pm 4.5$ & $100.0 \pm 0.0$ & $100.0 \pm 0.0$ \\
     & Few-shot & $73.3 \pm 7.1$ & $96.0 \pm 1.8$ & $100.0 \pm 0.0$ & $90.0 \pm 2.9$ & $100.0 \pm 0.0$ & $100.0 \pm 0.0$ & $76.4 \pm 5.3$ & $100.0 \pm 0.0$ & $100.0 \pm 0.0$ \\
     & Zero-shot+CoT & $95.3 \pm 3.2$ & $84.7 \pm 4.8$ & $100.0 \pm 0.0$ & $97.1 \pm 1.2$ & $100.0 \pm 0.0$ & $100.0 \pm 0.0$ & $62.0 \pm 6.1$ & $96.7 \pm 3.2$ & $100.0 \pm 0.0$ \\
     & Few-shot+CoT & $96.0 \pm 2.8$ & $94.7 \pm 2.1$ & $100.0 \pm 0.0$ & $87.9 \pm 3.4$ & $100.0 \pm 0.0$ & $100.0 \pm 0.0$ & $80.7 \pm 3.2$ & $100.0 \pm 0.0$ & $100.0 \pm 0.0$ \\
    \midrule
    \multirow{4}{*}{GPT-3.5} & Zero-shot & $84.7 \pm 5.6$ & $84.0 \pm 2.8$ & $100.0 \pm 0.0$ & $6.7 \pm 2.4$ & $64.0 \pm 9.9$ & $100.0 \pm 0.0$ & $0.0 \pm 0.0$ & $72.0 \pm 8.1$ & $100.0 \pm 0.0$ \\
     & Few-shot & $22.7 \pm 7.4$ & $86.0 \pm 5.6$ & $100.0 \pm 0.0$ & $8.7 \pm 3.6$ & $98.7 \pm 1.3$ & $96.7 \pm 2.6$ & $29.3 \pm 7.0$ & $91.3 \pm 6.0$ & $100.0 \pm 0.0$ \\
     & Zero-shot+CoT & $56.7 \pm 8.0$ & $88.0 \pm 3.3$ & $100.0 \pm 0.0$ & $2.7 \pm 2.6$ & $78.2 \pm 9.9$ & $100.0 \pm 0.0$ & $14.3 \pm 8.4$ & $81.4 \pm 6.7$ & $100.0 \pm 0.0$ \\
     & Few-shot+CoT & $57.5 \pm 12.1$ & $75.8 \pm 9.0$ & $100.0 \pm 0.0$ & $9.0 \pm 4.6$ & $96.0 \pm 3.8$ & $97.0 \pm 2.8$ & $34.3 \pm 8.0$ & $98.6 \pm 1.4$ & $100.0 \pm 0.0$ \\
    \midrule
    \multirow{4}{*}{LLama2} & Zero-shot & $0.7 \pm 0.7$ & $90.7 \pm 2.6$ & $100.0 \pm 0.0$ & $0.0 \pm 0.0$ & $89.2 \pm 3.8$ & $100.0 \pm 0.0$ & $0.0 \pm 0.0$ & $78.0 \pm 5.2$ & $100.0 \pm 0.0$ \\
     & Few-shot & $31.1 \pm 7.8$ & $100.0 \pm 0.0$ & $70.4 \pm 8.0$ & $17.2 \pm 5.0$ & $100.0 \pm 0.0$ & $82.8 \pm 5.0$ & $13.0 \pm 3.7$ & $100.0 \pm 0.0$ & $92.6 \pm 2.6$ \\
     & Zero-shot+CoT & $1.1 \pm 1.1$ & $90.4 \pm 1.9$ & $100.0 \pm 0.0$ & $0.0 \pm 0.0$ & $87.0 \pm 4.9$ & $100.0 \pm 0.0$ & $0.0 \pm 0.0$ & $70.9 \pm 5.0$ & $100.0 \pm 0.0$ \\
     & Few-shot+CoT & $16.2 \pm 5.2$ & $100.0 \pm 0.0$ & $87.2 \pm 5.2$ & $8.0 \pm 3.1$ & $100.0 \pm 0.0$ & $92.3 \pm 3.1$ & $13.3 \pm 3.7$ & $98.3 \pm 1.1$ & $91.7 \pm 3.3$ \\

    \bottomrule
  \end{tabular}
  }
\end{table*}

\begin{table*}
  \caption{The effect of different sampling temperatures on rule-based graph generation with GPT-4. ``Valid'' denotes fractions of generated graphs that are valid under specific rules, ``Unique'' denotes fractions of generated graphs that are not duplicates, and ``Novel`` denotes fractions of generated graphs that are different from the given example graphs.}
  \label{tab:rule-temperature}
  \centering
  \adjustbox{max width=\linewidth}{
  \begin{tabular}{cccccccccccc}
    \toprule
    \multirow{2}{*}{Temperature} &\multirow{2}{*}{Prompt}& \multicolumn{3}{c}{Cycles}& \multicolumn{3}{c}{$k$-regular}& \multicolumn{3}{c}{$k$-color}\\
    &&Valid&Unique&Novel&Valid&Unique&Novel&Valid&Unique&Novel\\
    \midrule

    \multirow{4}{*}{$t=0$} & Zero-shot & $100.0 \pm 0.0$ & $90.0 \pm 0.0$ & $100.0 \pm 0.0$ & $53.3 \pm 11.1$ & $98.7 \pm 0.9$ & $100.0 \pm 0.0$ & $33.3 \pm 6.3$ & $98.0 \pm 1.9$ & $100.0 \pm 0.0$ \\
     & Few-shot & $76.0 \pm 4.0$ & $89.3 \pm 1.8$ & $100.0 \pm 0.0$ & $90.8 \pm 3.0$ & $96.9 \pm 3.0$ & $100.0 \pm 0.0$ & $72.1 \pm 9.0$ & $90.7 \pm 4.0$ & $100.0 \pm 0.0$ \\
     & Zero-shot+CoT & $91.3 \pm 4.6$ & $88.7 \pm 5.6$ & $100.0 \pm 0.0$ & $80.0 \pm 10.1$ & $100.0 \pm 0.0$ & $100.0 \pm 0.0$ & $49.3 \pm 7.9$ & $96.4 \pm 2.8$ & $100.0 \pm 0.0$ \\
     & Few-shot+CoT & $96.7 \pm 3.2$ & $62.0 \pm 9.8$ & $100.0 \pm 0.0$ & $85.0 \pm 6.9$ & $100.0 \pm 0.0$ & $100.0 \pm 0.0$ & $52.7 \pm 8.9$ & $96.0 \pm 3.9$ & $100.0 \pm 0.0$ \\
    \midrule
    \multirow{4}{*}{$t=0.2$} & Zero-shot & $100.0 \pm 0.0$ & $90.0 \pm 0.0$ & $100.0 \pm 0.0$ & $34.7 \pm 9.9$ & $100.0 \pm 0.0$ & $100.0 \pm 0.0$ & $49.3 \pm 5.4$ & $100.0 \pm 0.0$ & $100.0 \pm 0.0$ \\
     & Few-shot & $79.3 \pm 4.5$ & $88.7 \pm 2.3$ & $100.0 \pm 0.0$ & $91.5 \pm 3.7$ & $100.0 \pm 0.0$ & $100.0 \pm 0.0$ & $70.0 \pm 8.1$ & $91.3 \pm 5.1$ & $100.0 \pm 0.0$ \\
     & Zero-shot+CoT & $87.3 \pm 4.8$ & $92.7 \pm 1.8$ & $100.0 \pm 0.0$ & $83.0 \pm 9.2$ & $98.0 \pm 1.9$ & $100.0 \pm 0.0$ & $50.7 \pm 9.1$ & $98.0 \pm 1.9$ & $100.0 \pm 0.0$ \\
     & Few-shot+CoT & $100.0 \pm 0.0$ & $77.3 \pm 8.8$ & $100.0 \pm 0.0$ & $91.3 \pm 2.8$ & $100.0 \pm 0.0$ & $100.0 \pm 0.0$ & $62.0 \pm 7.8$ & $88.7 \pm 6.5$ & $100.0 \pm 0.0$ \\
    \midrule
    \multirow{4}{*}{$t=0.4$} & Zero-shot & $100.0 \pm 0.0$ & $90.0 \pm 0.0$ & $100.0 \pm 0.0$ & $56.0 \pm 10.9$ & $99.3 \pm 0.6$ & $100.0 \pm 0.0$ & $52.0 \pm 7.0$ & $99.3 \pm 0.6$ & $100.0 \pm 0.0$ \\
     & Few-shot & $70.7 \pm 4.8$ & $84.0 \pm 2.8$ & $100.0 \pm 0.0$ & $89.2 \pm 4.1$ & $95.4 \pm 4.4$ & $100.0 \pm 0.0$ & $56.9 \pm 8.7$ & $86.2 \pm 7.6$ & $100.0 \pm 0.0$ \\
     & Zero-shot+CoT & $98.7 \pm 1.3$ & $84.7 \pm 5.2$ & $100.0 \pm 0.0$ & $76.0 \pm 9.1$ & $98.0 \pm 1.3$ & $100.0 \pm 0.0$ & $44.6 \pm 7.0$ & $94.6 \pm 3.7$ & $100.0 \pm 0.0$ \\
     & Few-shot+CoT & $97.9 \pm 2.1$ & $91.4 \pm 1.7$ & $100.0 \pm 0.0$ & $90.0 \pm 3.8$ & $100.0 \pm 0.0$ & $100.0 \pm 0.0$ & $58.7 \pm 10.6$ & $90.0 \pm 6.1$ & $100.0 \pm 0.0$ \\
    \midrule
    \multirow{4}{*}{$t=0.6$} & Zero-shot & $100.0 \pm 0.0$ & $90.7 \pm 1.1$ & $100.0 \pm 0.0$ & $50.7 \pm 10.9$ & $100.0 \pm 0.0$ & $100.0 \pm 0.0$ & $59.3 \pm 6.1$ & $90.7 \pm 5.8$ & $100.0 \pm 0.0$ \\
     & Few-shot & $77.3 \pm 6.4$ & $90.7 \pm 2.9$ & $100.0 \pm 0.0$ & $91.7 \pm 3.1$ & $100.0 \pm 0.0$ & $100.0 \pm 0.0$ & $64.0 \pm 8.2$ & $94.7 \pm 5.2$ & $100.0 \pm 0.0$ \\
     & Zero-shot+CoT & $90.0 \pm 5.7$ & $86.0 \pm 5.5$ & $100.0 \pm 0.0$ & $92.5 \pm 3.9$ & $100.0 \pm 0.0$ & $100.0 \pm 0.0$ & $45.7 \pm 6.8$ & $99.3 \pm 0.7$ & $100.0 \pm 0.0$ \\
     & Few-shot+CoT & $96.0 \pm 3.2$ & $75.3 \pm 8.6$ & $100.0 \pm 0.0$ & $82.9 \pm 6.6$ & $100.0 \pm 0.0$ & $100.0 \pm 0.0$ & $63.3 \pm 8.1$ & $100.0 \pm 0.0$ & $100.0 \pm 0.0$ \\
    \midrule
    \multirow{4}{*}{$t=0.8$} & Zero-shot & $91.3 \pm 3.3$ & $88.3 \pm 1.6$ & $100.0 \pm 0.0$ & $64.0 \pm 6.8$ & $100.0 \pm 0.0$ & $100.0 \pm 0.0$ & $50.3 \pm 5.5$ & $98.3 \pm 1.6$ & $100.0 \pm 0.0$ \\
     & Few-shot & $85.0 \pm 3.3$ & $92.3 \pm 1.3$ & $100.0 \pm 0.0$ & $86.1 \pm 3.1$ & $100.0 \pm 0.0$ & $100.0 \pm 0.0$ & $62.3 \pm 5.1$ & $98.5 \pm 1.5$ & $100.0 \pm 0.0$ \\
     & Zero-shot+CoT & $86.9 \pm 3.6$ & $89.0 \pm 4.2$ & $100.0 \pm 0.0$ & $82.7 \pm 8.6$ & $100.0 \pm 0.0$ & $100.0 \pm 0.0$ & $43.2 \pm 4.9$ & $98.6 \pm 1.0$ & $100.0 \pm 0.0$ \\
     & Few-shot+CoT & $97.0 \pm 1.9$ & $83.0 \pm 5.1$ & $100.0 \pm 0.0$ & $91.5 \pm 1.6$ & $100.0 \pm 0.0$ & $100.0 \pm 0.0$ & $58.5 \pm 5.9$ & $96.3 \pm 2.6$ & $100.0 \pm 0.0$ \\
    \midrule
    \multirow{4}{*}{$t=1$} & Zero-shot & $94.0 \pm 3.9$ & $88.0 \pm 3.0$ & $100.0 \pm 0.0$ & $60.0 \pm 10.5$ & $100.0 \pm 0.0$ & $100.0 \pm 0.0$ & $40.7 \pm 7.5$ & $100.0 \pm 0.0$ & $100.0 \pm 0.0$ \\
     & Few-shot & $64.0 \pm 6.2$ & $96.0 \pm 1.6$ & $100.0 \pm 0.0$ & $86.7 \pm 4.1$ & $100.0 \pm 0.0$ & $100.0 \pm 0.0$ & $49.2 \pm 8.9$ & $95.8 \pm 4.0$ & $100.0 \pm 0.0$ \\
     & Zero-shot+CoT & $95.0 \pm 2.2$ & $86.4 \pm 5.9$ & $100.0 \pm 0.0$ & $78.6 \pm 11.1$ & $98.6 \pm 1.3$ & $100.0 \pm 0.0$ & $57.1 \pm 6.9$ & $100.0 \pm 0.0$ & $100.0 \pm 0.0$ \\
     & Few-shot+CoT & $80.7 \pm 6.6$ & $91.3 \pm 2.5$ & $100.0 \pm 0.0$ & $81.5 \pm 4.2$ & $100.0 \pm 0.0$ & $100.0 \pm 0.0$ & $45.0 \pm 9.4$ & $100.0 \pm 0.0$ & $100.0 \pm 0.0$ \\

    \bottomrule
  \end{tabular}
  }
\end{table*}

\begin{table*}
  \caption{The effect of different amounts of generated graphs on rule-based graph generation with GPT-4. ``Valid'' denotes fractions of generated graphs that are valid under specific rules, ``Unique'' denotes fractions of generated graphs that are not duplicates, and ``Novel`` denotes fractions of generated graphs that are different from the given example graphs. Items marked with ``---'' denote experiments where the model is unable to generate enough graphs with the required format.}
  \label{tab:rule-ansnum}
  \centering
  \adjustbox{max width=\linewidth}{
  \begin{tabular}{cccccccccccc}
    \toprule
    \multirow{2}{*}{Amount} &\multirow{2}{*}{Prompt}& \multicolumn{3}{c}{Cycles}& \multicolumn{3}{c}{$k$-regular}& \multicolumn{3}{c}{$k$-color}\\
    &&Valid&Unique&Novel&Valid&Unique&Novel&Valid&Unique&Novel\\
    \midrule

    \multirow{4}{*}{5} & Zero-shot & $90.7 \pm 4.2$ & $81.3 \pm 2.3$ & $100.0 \pm 0.0$ & $57.3 \pm 11.1$ & $100.0 \pm 0.0$ & $100.0 \pm 0.0$ & $50.7 \pm 7.0$ & $100.0 \pm 0.0$ & $100.0 \pm 0.0$ \\
     & Few-shot & $72.0 \pm 4.1$ & $97.3 \pm 1.8$ & $100.0 \pm 0.0$ & $92.9 \pm 2.6$ & $100.0 \pm 0.0$ & $100.0 \pm 0.0$ & $69.3 \pm 5.3$ & $100.0 \pm 0.0$ & $100.0 \pm 0.0$ \\
     & Zero-shot+CoT & $86.7 \pm 4.5$ & $85.3 \pm 5.2$ & $100.0 \pm 0.0$ & $85.0 \pm 7.5$ & $100.0 \pm 0.0$ & $100.0 \pm 0.0$ & $44.0 \pm 4.7$ & $100.0 \pm 0.0$ & $100.0 \pm 0.0$ \\
     & Few-shot+CoT & $92.9 \pm 3.8$ & $82.9 \pm 7.2$ & $100.0 \pm 0.0$ & $93.3 \pm 3.1$ & $100.0 \pm 0.0$ & $100.0 \pm 0.0$ & $44.0 \pm 9.3$ & $100.0 \pm 0.0$ & $100.0 \pm 0.0$ \\
    \midrule
    \multirow{4}{*}{10} & Zero-shot & $91.3 \pm 3.3$ & $88.3 \pm 1.6$ & $100.0 \pm 0.0$ & $64.0 \pm 6.8$ & $100.0 \pm 0.0$ & $100.0 \pm 0.0$ & $50.3 \pm 5.5$ & $98.3 \pm 1.6$ & $100.0 \pm 0.0$ \\
     & Few-shot & $85.0 \pm 3.3$ & $92.3 \pm 1.3$ & $100.0 \pm 0.0$ & $86.1 \pm 3.1$ & $100.0 \pm 0.0$ & $100.0 \pm 0.0$ & $62.3 \pm 5.1$ & $98.5 \pm 1.5$ & $100.0 \pm 0.0$ \\
     & Zero-shot+CoT & $86.9 \pm 3.6$ & $89.0 \pm 4.2$ & $100.0 \pm 0.0$ & $82.7 \pm 8.6$ & $100.0 \pm 0.0$ & $100.0 \pm 0.0$ & $43.2 \pm 4.9$ & $98.6 \pm 1.0$ & $100.0 \pm 0.0$ \\
     & Few-shot+CoT & $97.0 \pm 1.9$ & $83.0 \pm 5.1$ & $100.0 \pm 0.0$ & $91.5 \pm 1.6$ & $100.0 \pm 0.0$ & $100.0 \pm 0.0$ & $58.5 \pm 5.9$ & $96.3 \pm 2.6$ & $100.0 \pm 0.0$ \\
    \midrule
    \multirow{4}{*}{20} & Zero-shot & $98.2 \pm 1.1$ & $92.1 \pm 1.6$ & $100.0 \pm 0.0$ & $53.7 \pm 10.1$ & $99.0 \pm 0.5$ & $100.0 \pm 0.0$ & $56.7 \pm 5.9$ & $92.3 \pm 4.9$ & $100.0 \pm 0.0$ \\
     & Few-shot & $83.7 \pm 4.7$ & $88.0 \pm 2.9$ & $100.0 \pm 0.0$ & $93.0 \pm 1.6$ & $100.0 \pm 0.0$ & $100.0 \pm 0.0$ & $65.0 \pm 8.3$ & $96.8 \pm 3.1$ & $100.0 \pm 0.0$ \\
     & Zero-shot+CoT & $95.0 \pm 3.2$ & $85.4 \pm 6.9$ & $100.0 \pm 0.0$ & --- & --- & --- & --- & --- & --- \\
     & Few-shot+CoT & $96.3 \pm 2.0$ & $93.7 \pm 2.2$ & $100.0 \pm 0.0$ & $95.0 \pm 1.4$ & $100.0 \pm 0.0$ & $100.0 \pm 0.0$ & $80.0 \pm 6.2$ & $82.9 \pm 8.6$ & $100.0 \pm 0.0$ \\

    \bottomrule
  \end{tabular}
  }
\end{table*}

\end{document}